\documentclass[letterpaper, 10 pt, conference]{ieeeconf}  % Comment this line out if you need a4paper

\usepackage{comment}
\usepackage{amsmath,amssymb}
\usepackage{color}
\usepackage{url}
\usepackage{hyperref}
\usepackage{graphicx}
\usepackage{amsmath,amssymb}
\usepackage{tikz}
\usepackage{cite}
\usepackage{floatrow}
\usepackage{comment}
\usepackage{amsmath,amssymb} % define this before the line numbering.
\usepackage{color}
\usepackage{floatrow}
\usepackage{array}
\usepackage{float}
\floatstyle{plaintop}

\restylefloat{table}
\newcolumntype{M}{>{\centering\arraybackslash}m{\dimexpr.22\linewidth-2\tabcolsep}}
\newcolumntype{S}{>{\centering\arraybackslash}m{\dimexpr.05\linewidth-5\tabcolsep}}
\DeclareMathOperator{\sign}{sgn}

\IEEEoverridecommandlockouts                              % This command is only needed if 
\title{\LARGE \bf
Fast Region of Interest Proposals on Maritime UAVs
}

\author{Benjamin Kiefer$^{1}$ and Andreas Zell$^{1}$% <-this % stops a space
\thanks{*This work was supported by the German Ministry
for Economic Affairs and Energy, Project Avalon, FKZ:
03SX481B.}% <-this % stops a space
\thanks{$^{1}$Both authors are with the Faculty of Computer Science,
        University of Tuebingen, Germany.
        {\tt\small prename.surname@uni-tuebingen.de}}%
}

\begin{document}

\maketitle
\thispagestyle{empty}
\pagestyle{empty}

%%%%%%%%%%%%%%%%%%%%%%%%%%%%%%%%%%%%%%%%%%%%%%%%%%%%%%%%%%%%%%%%%%%%%%%%%%%%%%%%
\begin{abstract}

Unmanned aerial vehicles assist in maritime search and rescue missions by flying over large search areas to autonomously search for objects or people. Reliably detecting objects of interest requires fast models to employ on embedded hardware. Moreover, with increasing distance to the ground station only part of the video data can be transmitted. In this work, we consider the problem of finding meaningful region of interest proposals in a video stream on an embedded GPU. Current object or anomaly detectors are not suitable due to their slow speed, especially on limited hardware and for large image resolutions. Lastly, objects of interest, such as pieces of wreckage, are often not known a priori. Therefore, we propose an end-to-end future frame prediction model running in real-time on embedded GPUs to generate region proposals. We analyze its performance on large-scale maritime data sets and demonstrate its benefits over traditional and modern methods.

\end{abstract}

%%%%%%%%%%%%%%%%%%%%%%%%%%%%%%%%%%%%%%%%%%%%%%%%%%%%%%%%%%%%%%%%%%%%%%%%%%%%%%%%

\section{Introduction}

Autonomous vision aboard UAVs has grown to an
important research area \cite{zhu2018visdrone,menouar2017uav,lygouras2019unsupervised,mishra2020drone}. Next to traffic surveillance \cite{fan2020visdrone,du2018unmanned}
and agriculture \cite{tsouros2019review}, also the field of search and rescue (SaR) has been
tackled \cite{varga2021seadronessee,mishra2020drone}. However, while several works focus on path-planning and mission implementation \cite{bevacqua2015mixed,hayat2020multi,mayer2019drones}, few works address the actual vision part, necessary for autonomously searching certain
areas.

Finding interesting regions on the sea is a hard problem,
since objects of interest are often not known a priori or
have a vast variety of different appearances which is why
supervised methods often fail in these scenarios. Even if object categories are
known beforehand, current methods focus on object detection,
which is not viable for large image resolutions and real-time (for rigor defined here to be $>$25FPS) performance on embedded hardware \cite{cazzato2020survey}. Both constraints occur in reliable SaR missions.

Furthermore, labeled data sets in these environments are
scarce as the data acquisition process is a complicated undertaking, requiring strict safety regulations for all subjects, and is expensive \cite{seadronesseedataacqui}. Instead, it is considerably easier
and cheaper to obtain raw data of sea surfaces.

What is more, often a low bandwidth, possibly due to large
distances or suboptimal weather conditions, does not allow for
the whole footage being transmitted to a ground station. This becomes especially severe in maritime scenarios, where the drone is far away from any ground station \cite{avalon,larus}. While compression
can be done on-board, it is often not sufficient and furthermore
results in image quality loss across the whole image, i.e. also
possibly quality loss in regions of the image that need to
be analyzed more thoroughly to exclude false positives or
negatives.

Recently, special purpose video codecs that allow few regions of an image to be coded with near constant picture quality have been proposed to tackle this problem \cite{steinert2022architecture}. The high quality regions that are transmitted can subsequently be combined with an actual classical object detection system on a ground station with much more hardware resources.

This methodology separates the problem into two stages: generating few high-recall regions of interest of a high dimensional image in real-time in a low resource environment and classifying these regions into known classes on a ground station with more resources. Motivated by these observations, in this work, we formulate and formalize the former problem and propose an autoencoder-based future frame prediction model that generates meaningful regions of interest on sea surfaces which can run in real-time on an embedded GPU. Owing to the nature of maritime environments, we show that classical methods perform poorly due to dynamic backgrounds, wave movements, sun reflections and others while modern methods are too slow. As this method is a type of anomaly detection method, it does not require bounding box annotations. We introduce a metric that measures the recall at a given amount of footage being transmitted and show that this method outperforms classical methods on multiple benchmarks and metrics.

To train the proposed model, we capture over 60 minutes of 4K video footage with several cameras depicting the sea surface from different angles and altitudes at different days and waters. We provide a webserver where we host the Maritime Anomaly Detection Benchmark, where researchers can upload their predictions, which will be evaluated on the server side to allow for a fair comparison.

\begin{figure*}
\begin{tabular}{c}
\centering
			% trim: left, bottom, right, top
		%\includegraphics[trim=0 0 0 %0,clip,width=.9\textwidth]{images/autoencoder_sketch.png}
		\includegraphics[trim=0 0 0 0,clip,width=.9\textwidth]{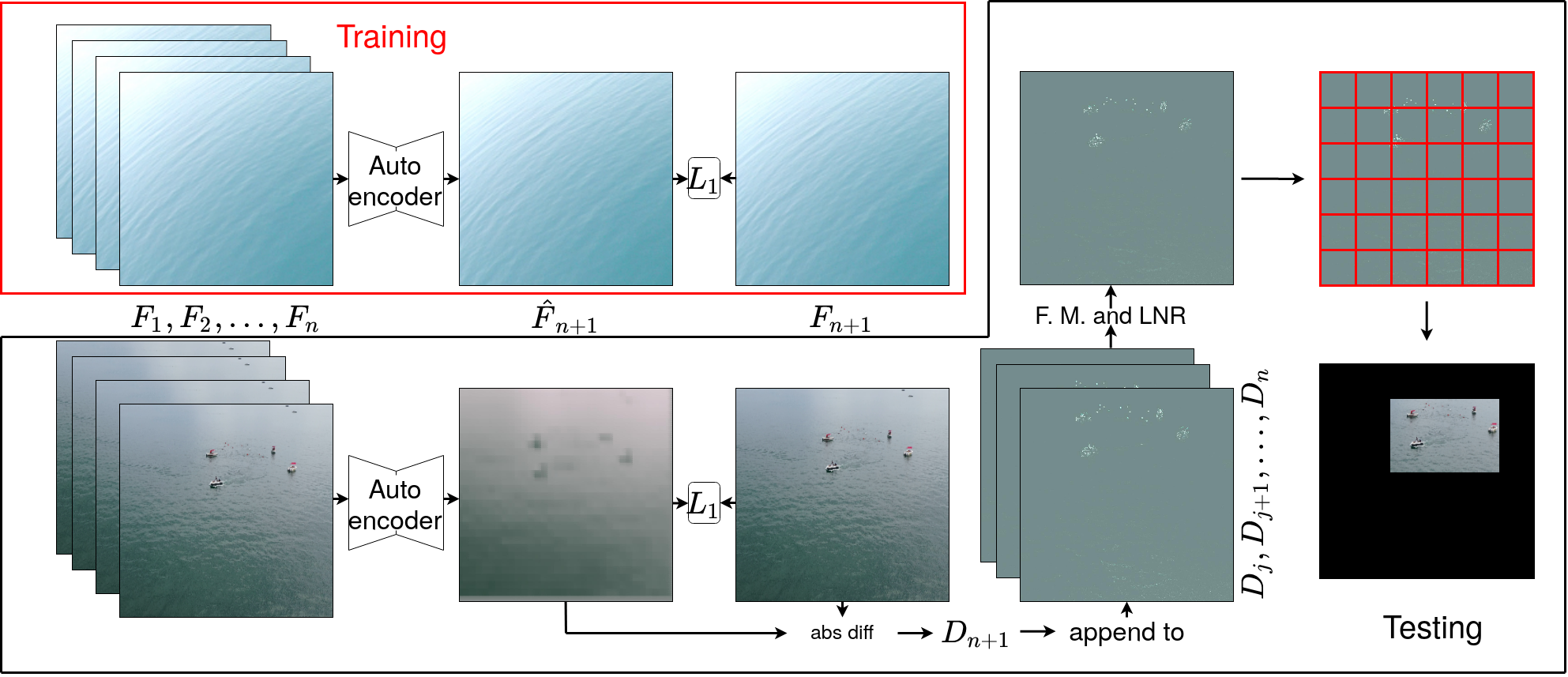}

\end{tabular}

\caption{Best viewed digitally. Future frame prediction autoencoder pipeline. The frames $F_1,\dots,F_n$ are concatenated and input into the autoencoder, which learns to predict $F_{n+1}$ via $\hat{F}_{n+1}$ and the $L_1$ loss. The error frame (absolute difference between the two), $D_{n+1}$, is concatenated to the last $D_j,\dots , D_n$. Then ,frame momentum and local noise reducer are applied until a final grid is put on the resulting error frame to yield final regions of interests. }
	\label{fig:autoencoder_img}
\end{figure*}

Our contributions are as follows:
\begin{itemize}

	\item We formulate a novel problem of obtaining high-recall regions of interest in a high-resolution and real-time scenario and propose a future frame prediction autoencoder to detect these regions in real-time on an embedded GPU.
	
	\item We capture over 60 minutes of video footage of the sea surface in various conditions as train set for our method and make it publicly available\footnote{\url{https://seadronessee.cs.uni-tuebingen.de/}}. We host a web server and propose the Maritime Anomaly Detection Benchmark with upload options.
	
	\item We analyze the proposed method and compare it to traditional and modern methods on two large-scale public data sets. 
	
\end{itemize}

\subsection{Related Work}

%\subsubsection{Airborne Maritime Data Sets}
\subsubsection{Computer Vision in Maritime Environments}

Airborne maritime data sets are scarce and mostly focus on synthetic aperture radar satellite imagery and ships \cite{airbus-ship,chen2020fgsd,wang2019sar,zhang2021sar}. \cite{marques2015unmanned,varga2021seadronessee,lygouras2019unsupervised} provide UAV-based maritime detection data sets. While the data set in \cite{lygouras2019unsupervised} features only stock photos scraped from the internet, the Seagull data set \cite{marques2015unmanned} and SeaDronesSee \cite{varga2021seadronessee} provide video material with objects of interest. Of these data sets, only Seagull provides frames that do not contain objects. However, the videos suffer from heavy lens distortion and distortion caused by a rolling shutter. We collect 60 minutes of video footage ($>$100000 frames) depicting the sea surface in various altitudes at different angles and days with multiple cameras. We weakly annotate the footage such that no objects of interest are visible in any of the frames.
\subsubsection{UAV-based Detection}
Rudimentary vision methods in SaR scenarios aboard a UAV are done in \cite{scherer2015autonomous}, using color, text or shape cues using OpenCV \cite{bradski2000opencv} to detect objects of interest. Similarly, \cite{rudol2008human} use Haar features to detect objects of interest in SaR missions. Among the learning-based methods \cite{ferreira2020ship} consider the application of ship detection, classifying images into positives (containing ships) and negatives. While it is also an unsupervised method, they ignore the localization. \cite{lygouras2019unsupervised} describe a complete SaR system from path planning over detection to action. However, their detection system is a basic YOLO variant unsuitable for large resolutions and real-time. Furthermore, it is restricted to the objects it is trained on. Generally, all literature regarding supervised UAV object detection can be considered related \cite{fan2020visdrone,zhu2018visdrone,du2018unmanned,varga2021seadronessee,xia2018dota,kiefer2021leveraging,price2018deep}, albeit not viable, since they do not work in real-time large-resolution scenarios on embedded hardware.

\subsubsection{Region proposal networks}

Selective search \cite{uijlings2013selective,van2011segmentation} generates many thousand little informative boxes for use of an object detector at a later stage, which makes it inapplicable in embedded environments. Common region proposal networks \cite{zhong2019anchor} are used in two-stage object detectors, such as Faster R-CNN \cite{ren2015faster}, but require bounding box supervision to be learned. Methods for weakly supervised object detection often employ region proposal networks \cite{tang2018weakly}, but require image-level annotations.

Background subtraction methods \cite{benezeth2010comparative} are used to separate the background from the foreground, which is defined by the scene captured by a static camera. Most of the methods are not suitable for dynamically changing scenes caused by camera and background movement. Furthermore, these methods do not focus on obtaining meaningful bounding box locations but only on the obtained segmentation maps.

(Video) Anomaly detection methods \cite{deecke2018image,sultani2018real,nguyen2019anomaly} learn on normal samples and detect anything previously not seen as anomalies. In images, this is often done for industrial parts \cite{staar2019anomaly,roth2021towards}, and in static videos for surveillance in traffic and crowded scenes \cite{saligrama2012video,zhao2017spatio,zhou2019anomalynet,liu2018future}. Earlier methods only focus on classifying images or frames \cite{ferreira2020ship,liu2018future}, while newer methods also consider localizing anomalies \cite{li2021cutpaste,Szymanowicz_2022_WACV}. However, these methods either ignore the temporal dimension or are not suitable for real-time use. Furthermore, video anomaly methods are designed for static scenes. Our work focuses on dynamic scenes and requires models running in real-time on embedded hardware. Furthermore, the focus is on generating high recall meaningful bounding box regions that potentially contain objects of interest. 

%We incorporate ideas from background subtraction while using methods based on video anomaly detection with a different aim as region proposal networks.

%subsection for maritime environments, what are the challenges here?

\section{Method}

We are given a high resolution (e.g. 4K) video stream depicting the sea surface. Furthermore, we have an embedded GPU (e.g. Nvidia Xavier AGX). The task is to select regions of interest in every frame which are to be transmitted down (e.g. via a streaming FPGA \cite{steinert2022architecture}). Each region is defined via four bounding box locations, describing the corners of the region in pixels (similar to classical object detection). Depending on the exact use case, the remaining regions are either also transmitted with lower quality or completely omitted.

\begin{figure}
	\centering

		\includegraphics[trim=0 0 0 0,clip,width=.9\textwidth]{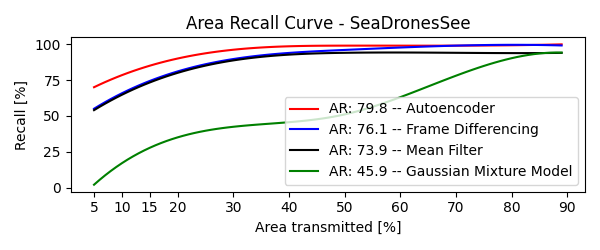}\\
		\includegraphics[trim=0 0 0 0,clip,width=.9\textwidth]{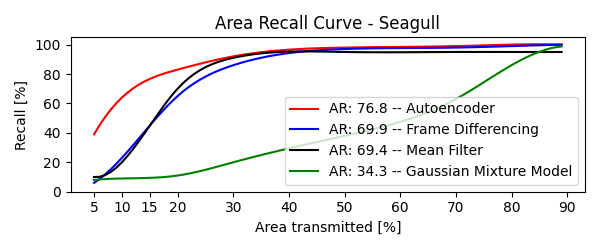}

	\caption{Area recall curves for SeaDronesSee and Seagull. AR$\hat{=}$avg. recall.}

	\label{fig:arearecall}
\end{figure}

We propose an autoencoder-based future frame prediction architecture to detect anomalies (See Fig. \ref{fig:autoencoder_img}). We train a shallow autoencoder on sequences of normal images $F_1,\dots,F_n$ depicting the sea surface such that the model learns to predict the next normal frame. Subsequently, the predicted frame $\hat{F}_{n+1}$ is subtracted from the original next frame $F_{n+1}$. The hypothesis is that the autoencoder fails to reconstruct objects that differ from the sea surface in their colors, shapes and textures (e.g. see Fig. \ref{fig:arearecall} \& \ref{fig:manyimgs}). Furthermore, by incorporating the last few frames, the autoencoder learns temporal correlations of water movements. 

%As we work with high resolutions and in real-time, we employ a very shallow autoencoder architecture having as low as six convolutional layers. To account for the low capacity of these autoencoders, we leverage freely available meta data to propose a codebook for autoencoders, denoted Autoencodebook. We divide the space of all available meta data into bins of certain sizes to increase the capacity without sacrificing inference speed. In Section \ref{sec:experiment}, we test various autoencoder architectures and examine the influence of different settings. 

Common future frame prediction networks employ large networks, such as UNet \cite{liu2018future,szymanowicz2022discrete} or even larger models \cite{yu2020efficient}. They operate on small video resolutions and are not suitable for employment on embedded hardware. Applying models in real-time on embedded hardware and for high resolutions requires us to fall back to shallow autoencoder architectures. We follow the basic principle of an encoder-decoder architecture, but only employ small channel dimensions for the filters as these make up for a large computational overhead. We refrain from using depth-wise separable convolutions \cite{howard2017mobilenets} or more advanced methods \cite{lebedev2014speeding}, since they are not optimized for embedded GPUs. For the first layer, we concatenate the past $n$ frames along the channel dimension and apply a regular 2D convolution with filter dimension $n\times 4$, kernel size $3\times 3$ and stride $2$. We perform the same convolution six times, while halving the channel dimension each time due to performance. The decoder performs the symmetric operations via deconvolutions.

\begin{figure}

% trim: left, bottom, right, top
   \centering

 \setlength{\tabcolsep}{1pt}
	\begin{tabular}{ccc}
			% trim: left, bottom, right, top

	   (a) Raw & (b) no momentum & (c) momentum\\
		\includegraphics[trim=0 0 0 0,clip,width=.32\textwidth]{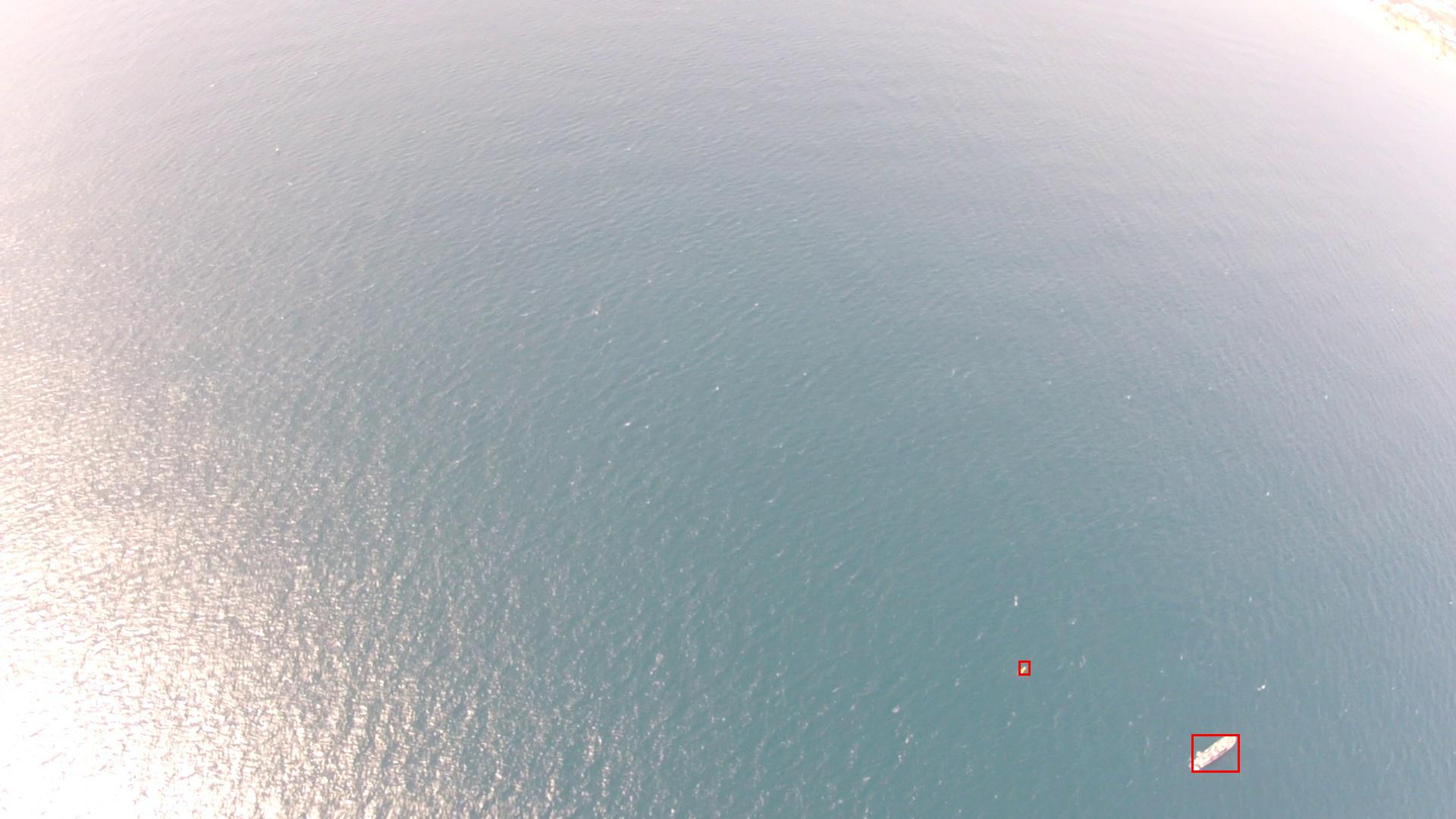} &
       \includegraphics[trim=0 0 0 25,clip,width=.32\textwidth]{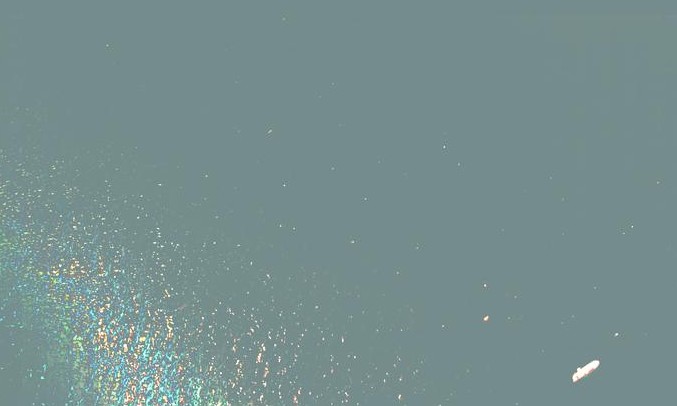} &
       \includegraphics[trim=0 0 0 25,clip,width=.32\textwidth]{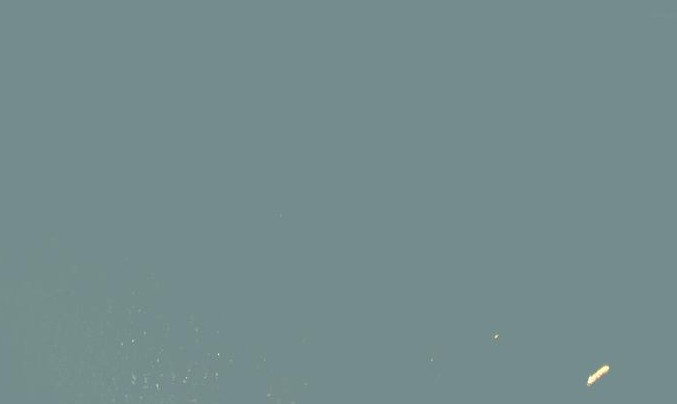}\\
     (d) Raw cut-out & (e) no LNR & (f) LNR\\
       \includegraphics[trim=0 0 0 1,clip,width=.32\textwidth]{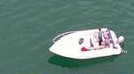}&
       \includegraphics[width=.32\textwidth]{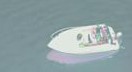}&
       \includegraphics[width=.32\textwidth]{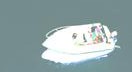}
       \end{tabular}
   
        \caption{Qualitative errors on Seagull (top) and SeaDronesSee (bottom). Note that the random noise at the bottom left of (c) almost vanished.}

 \label{fig:localnoise}

 \label{fig:frame-moment}
\end{figure}

We hypothesize that reconstruction errors due to wave patterns and sun reflections are more temporally unstable than actual anomalies. Thus, we propose to include an error frame momentum term, which averages over the past $n$ error frames $D_1,...,D_n$. This assumes that the camera movement is not too quick as then, the actual anomalies also move quickly in the image plane, eliminating the error frame momentum effect. However, for frame rates of roughly 30, this is negligible. Figure \ref{fig:frame-moment} (c) and Section \ref{sec:experiment} show the advantage of using this component.

To counteract the local noise induced by an imperfect reconstruction coming from sun reflections and wave patterns, we introduce a local noise remover (LNR). Channel-wise, we multiply each pixel of the error frame by its immediate vertical and horizontal surrounding neighbour. We repeat this procedure three times. This ensures that only regions of larger error areas are detected as anomalies (as opposed to noisy areas) in subsequent steps. This can be seen as a morphological operation, however also different, since we do not use a structuring element \cite{zhuang1986morphological}. See how the waves are eliminated in Fig. \ref{fig:localnoise} (f), while the boat is amplified.

\begin{figure}

	\begin{tabular}{@{}c@{\hspace{.99cm}}c@{}}
			% trim: left, bottom, right, top
		\includegraphics[trim=0 20 0 0,clip,width=.99\textwidth]{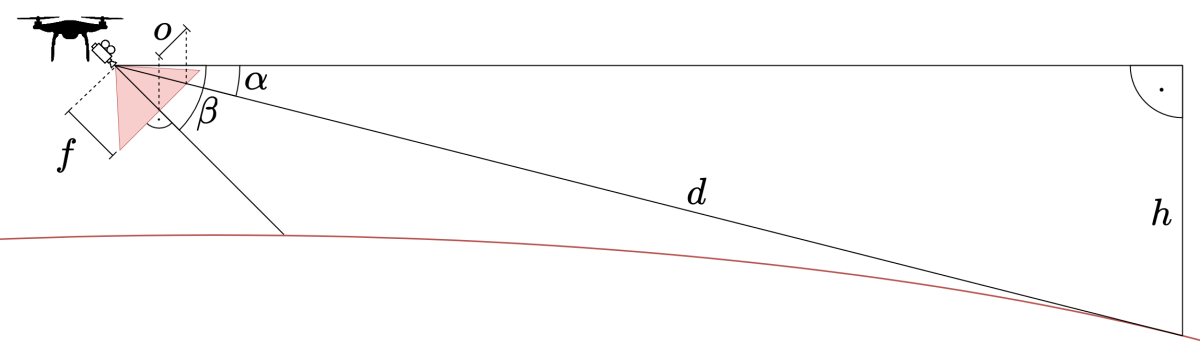}\\
		\includegraphics[trim=0 0 0 0,clip,width=.49\textwidth]{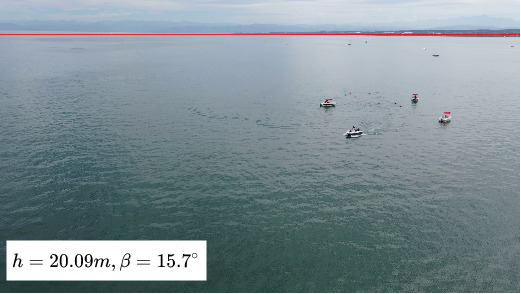}
		\includegraphics[trim=0 0 0 0,clip,width=.49\textwidth]{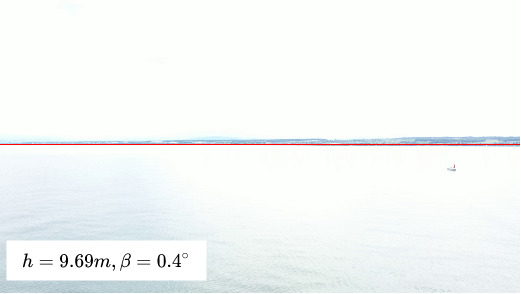}
		
	\end{tabular}
	\caption{Illustration of the horizon cutter (top) and predictions (bottom). The curvature is just for visualization purpose (ignored for computation).}
	\label{fig:horizon_cutter}
\end{figure}

%finalDO: gesoindert nochmal erwähnen, dass die challenges der übertragung in maritimen szenarien vorkommen!

Importantly, this method is sensitive to regions above the horizon. Therefore, we leverage meta data from the UAV’s on-board sensors to determine the horizon line in open
water. This allows us to ignore this region in the autoencoder training and inference phase
which, in turn, results in more robust anomaly detection performance and faster inference
times. Notably, this computation has virtually no overhead.
The horizon line can be computed using the UAV’s height, camera gimbal pitch and
roll angle, and the camera intrinsics. Ignoring the effect of atmospheric refraction, we can
estimate the distance to the horizon as a function of the height of the observer as $d=3.57h^{1/2}$. This approximation is fairly accurate for heights that are typical for SaR-UAVs (far below 1000m) \cite{bohren1986altitude}. We furthermore ignore the curvature of the earth, which is also negligible for these heights. We compute the angle $\alpha$ to the horizon via $\alpha=\arcsin (h/d)$. Using the focal length $f$ (in pixels) and the camera gimbal pitch $\beta$, we can then compute the camera perspective projection on the image plane, which yields the height offset $o$ in pixels to the horizontal center line of the image plane as $o=\tan (|\alpha-\beta|) \cdot f \cdot \sign (\alpha-\beta)$. Naturally, we truncate $o$ to be within the range of the number of horizontal pixels. To account for the roll angle $\gamma$ of the UAV (or camera gimbal), we can simply add a roll angle
induced offset at the left and subtract at the right of the image given as $o_r=\tan (\gamma)\cdot p_w/2$, where $p_w$ is the pixel width of the video. While the horizontal pixel location o is an approximation, it is quite robust to errors in the altitude h. Since an exact error analysis is not in the scope of this work, we just report values
for altitudes that are common in the data set SeaDronesSee. For $\beta = 0 - 20^\circ, h < 300m$ it holds that $10m$ in altitude error results in approx. $1px$ offset change in a 4K image. However, $o$ is very sensitive to errors in the gimbal angle $\beta$ . For example, for $h = 130m, \beta = 16^\circ$, $1^\circ$ in angle error results in approx. $40px$ offset change. Therefore, it is essential to have a well-calibrated gimbal and UAV IMU. The latter can be accurate up to $0.1^\circ$ when configured properly \cite{suzuki2016precise}. See Figure \ref{fig:horizon_cutter} for an illustration.
Empirically, we show that the horizon cutter performs well despite the occurrence of
nearby land. We manually annotate the horizon line for a subset of the SeaDronesSee-Tracking validation set and compute the pixel offset error and the roll angle $\gamma$ error. Table \ref{table:empiricalhorizon} shows that despite some land mass blocking the horizon (also see Fig. \ref{fig:horizon_cutter}), the error is negligible.

\begin{table}

	\caption{Error to the ground truth horizon as measured by roll angle $\gamma$ and pixel offset $o$ in a 4K image.}
	\begin{tabular}{c|c|c|c}
	Errors of &	$\gamma$  & $o$ &  $o/2160$ \\
		\hline
	horizon visible (5\%) & $0.8^\circ$ & $71$px & $3.3\%$\\
	horizon not visible (95\%) & -- & $2$px & $0.1\%$\\
	total & -- & $5.45$px & 0.3\% \\
	\end{tabular}

	\label{table:empiricalhorizon}
\end{table}

Lastly, we apply a grid of size $m_1\times m_2$ on the error frame and for every grid window we average over the error frame pixels contained in it. We select the maximum number of boxes outputted given a certain bandwidth, which we simply break down in $p\%$ of the area of the whole frame.

\section{Data Set Generation and Webserver}

To test our approach, we gathered 60 minutes of 4K video footage on open water at three different days with three different cameras. We made sure to include altitudes and viewing angles from $5-120m$ and $0-90^\circ$. We manually filtered out the sequences that contained objects considered anomalous, such as humans, boats, life jackets and buoys. Each frame is annotated with its corresponding meta data information, such as altitude, all angles of the UAV principal axes, camera gimbal pitch angle, time, GPS and others.

%finalDO rebuttal: change this to seadronessee

This data, called {\bf OpenWater}, comes along with over 20 minutes of bounding box annotated footage in open water where we annotated the same classes as in SeaDronesSee, serving as anomalies. Everything but the test annotations will be uploaded to avoid researchers from overfitting. We will upload the test videos, on a web server and propose the Weakly Supervised Maritime Anomaly Detection Benchmark where researchers can upload their predictions, which will be evaluated and published on the server side for fair comparisons.

\section{Experiments}
\label{sec:experiment}

For our experiments, we choose a grid size of $48 \times 27$, predict the fifth frame from the past four, and use an error frame momentum of two. Influences of the components are discussed in further sections.

%glaub oben wie unten nur mit cv2
%no cv2
%--savesampleimgs
%False
%--diffmomentum
%2
%--overlap
%0.3
%--gridmode
%True
%%--maxpercentmode
%0.05
%--usecv2contours
%False
%--horizoncutter
%True
%--autoencodebook
%False

%current recall of Autoencoder  :  .8611
%current recall of Naive Baseline  :  .7172
%current recall of Frame Differencing  :  0.6778
%Gaussian Mixture  :  0.5721

\begin{figure*}
	\centering

	\begin{tabular}{@{\hspace{.45cm}}c@{\hspace{.45cm}}c@{}}
			% trim: left, bottom, right, top
		\rotatebox{90}{\ \ Image}  \includegraphics[trim=0 0 0 0,clip,width=.22\textwidth]{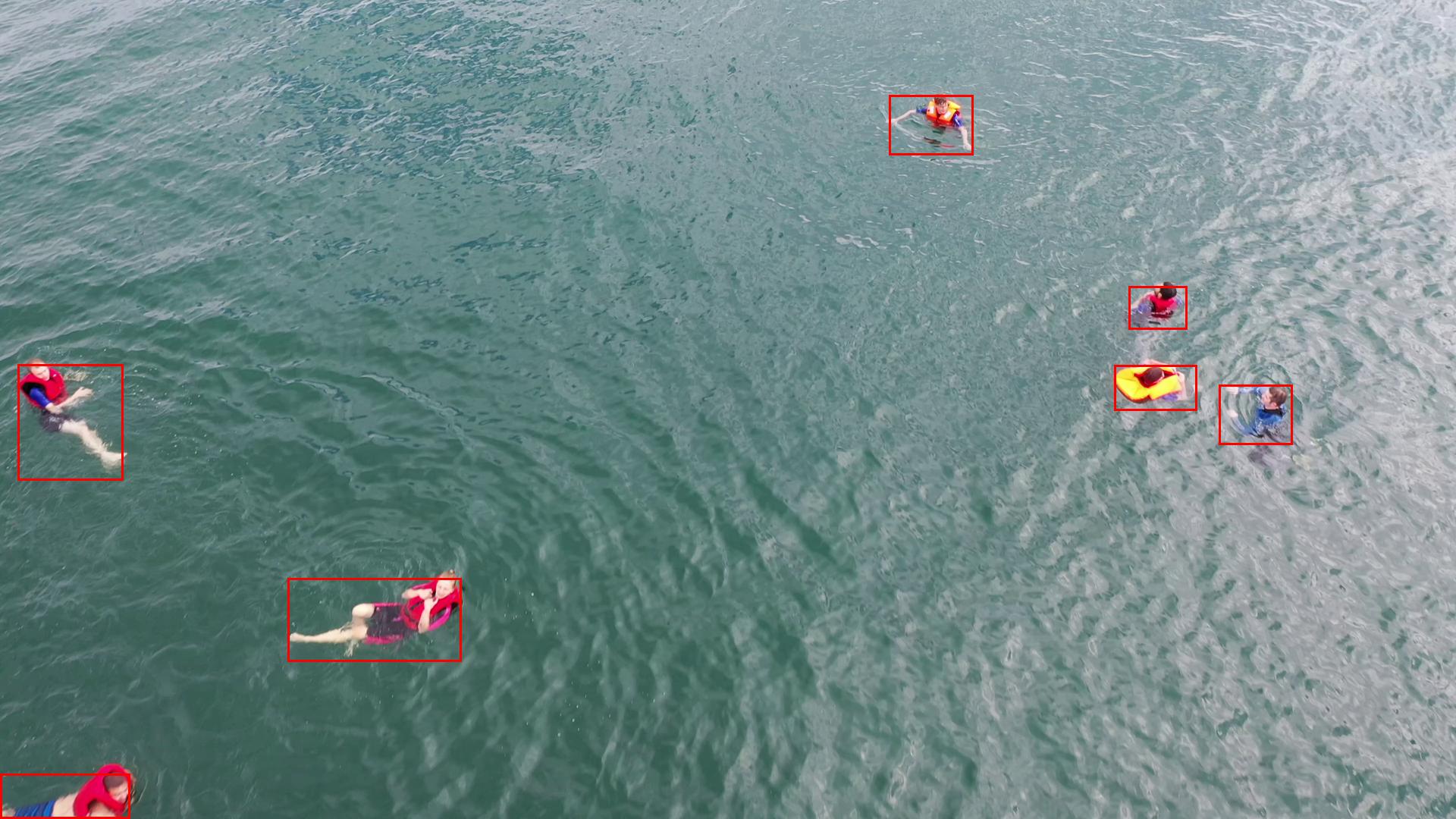}
		
		\includegraphics[trim=0 0 0 0,clip,width=.22\textwidth]{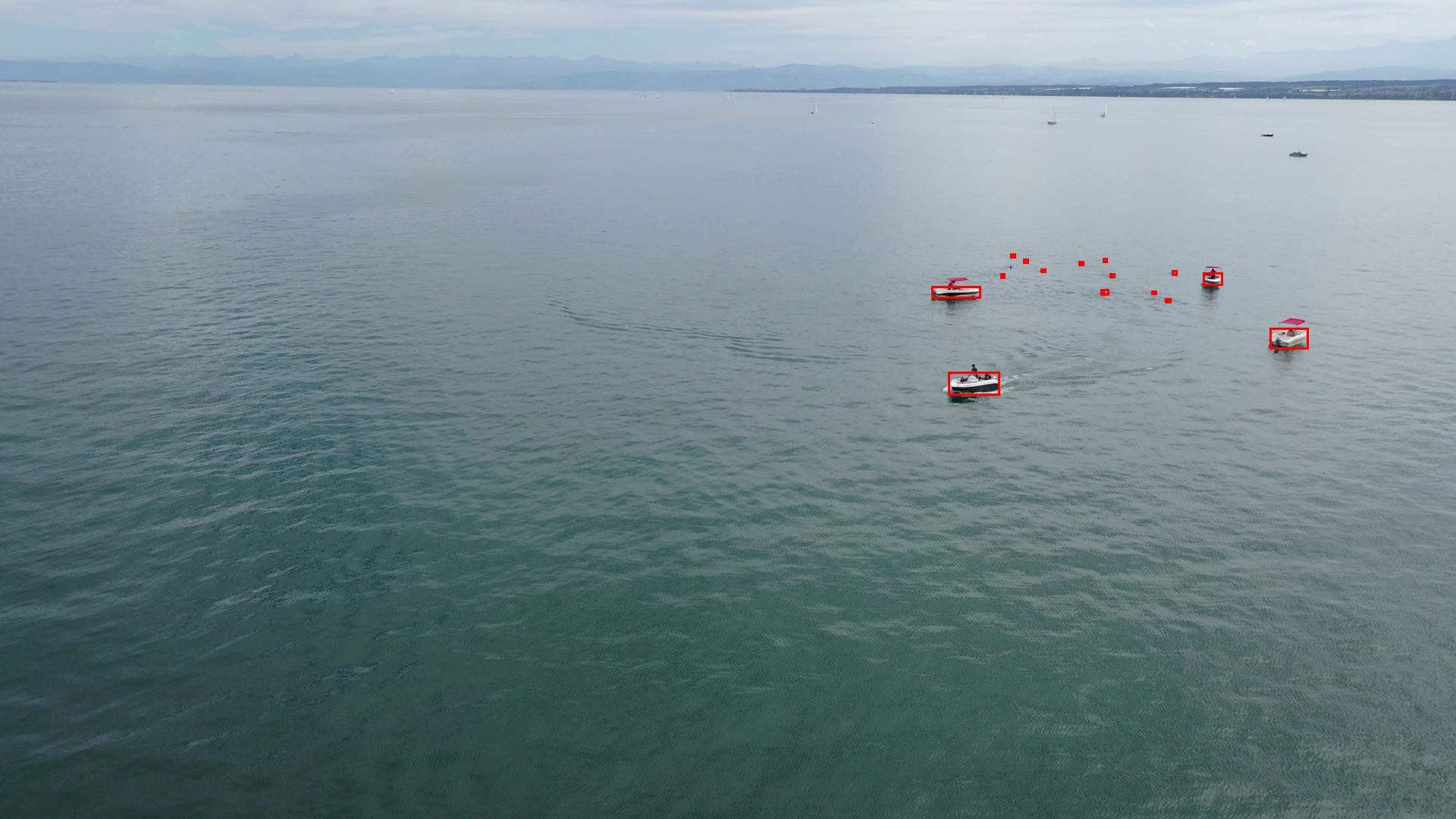}
		
		\includegraphics[trim=0 0 0 0,clip,width=.22\textwidth]{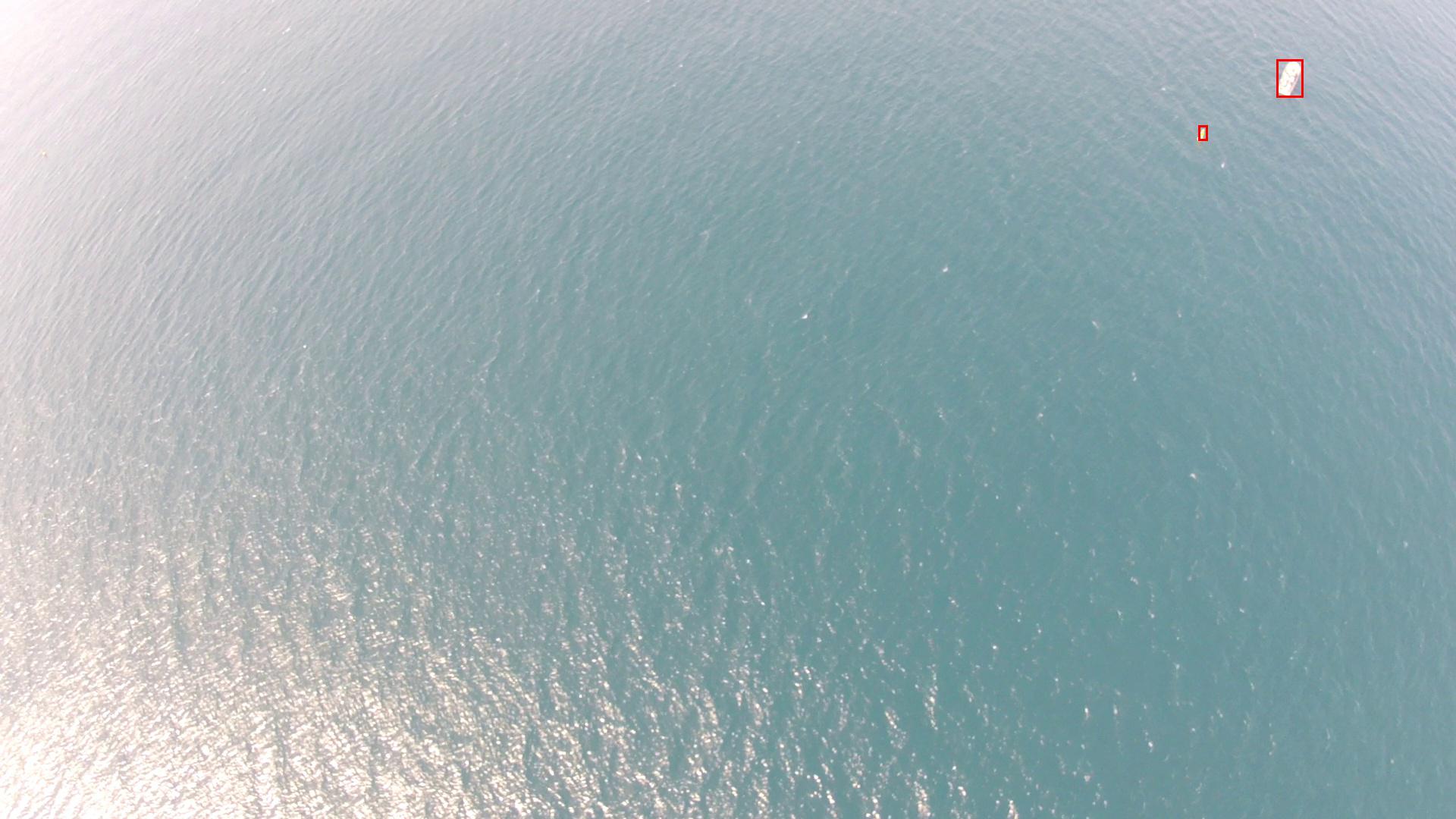}
		
		\includegraphics[trim=0 0 0 0,clip,width=.22\textwidth]{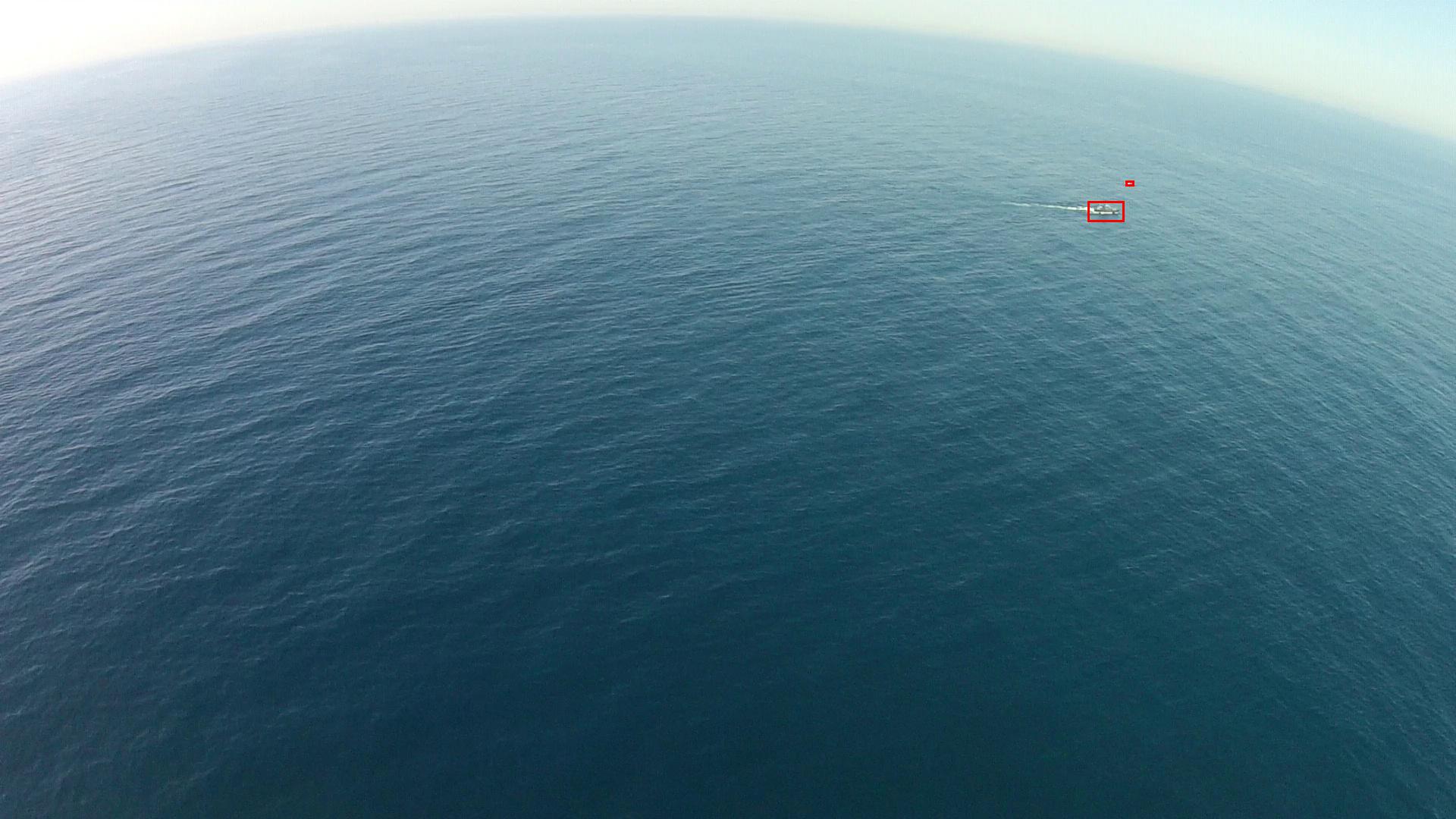}
		
		\\
		
		%\rotatebox{90}{MF err} \includegraphics[trim=0 0 0 0,clip,width=.22\textwidth]{images/img_Naive_Baseline5.jpg}
		
		%\includegraphics[trim=0 0 0 0,clip,width=.22\textwidth]{images/img_Naive_Baseline1276.jpg}
		
		%\includegraphics[trim=0 0 0 0,clip,width=.22\textwidth]{images/img_Naive_Baselineframe274.jpg}
		
		%\includegraphics[trim=0 0 0 0,clip,width=.22\textwidth]{images/img_Naive_Baselineframe136.jpg}
		%\\
		
		\rotatebox{90}{\ \ MF box}  \includegraphics[trim=0 0 0 0,clip,width=.22\textwidth]{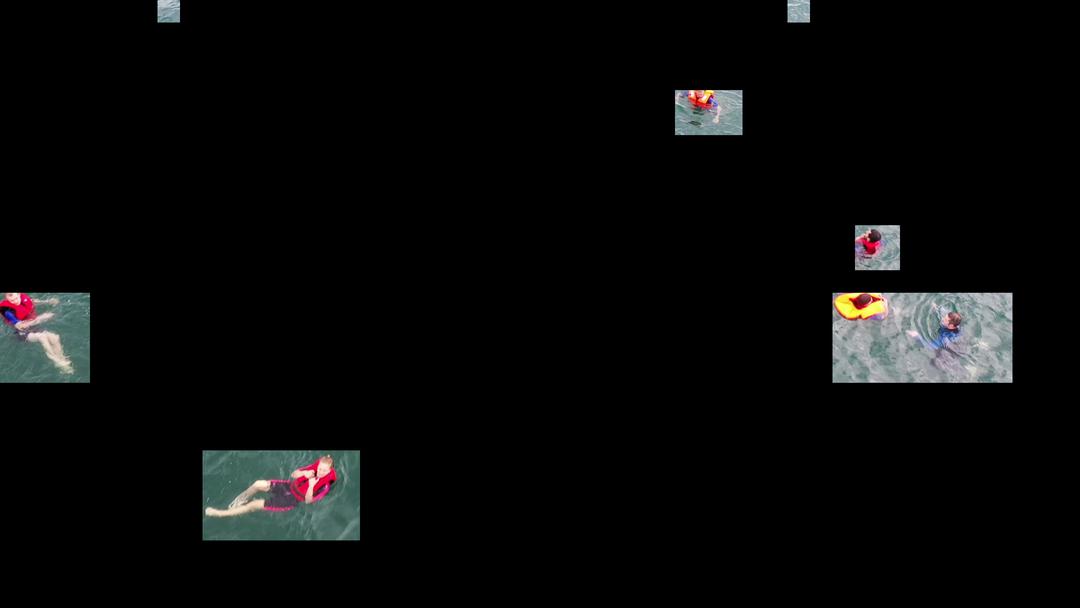}
		
		\includegraphics[trim=0 0 0 0,clip,width=.22\textwidth]{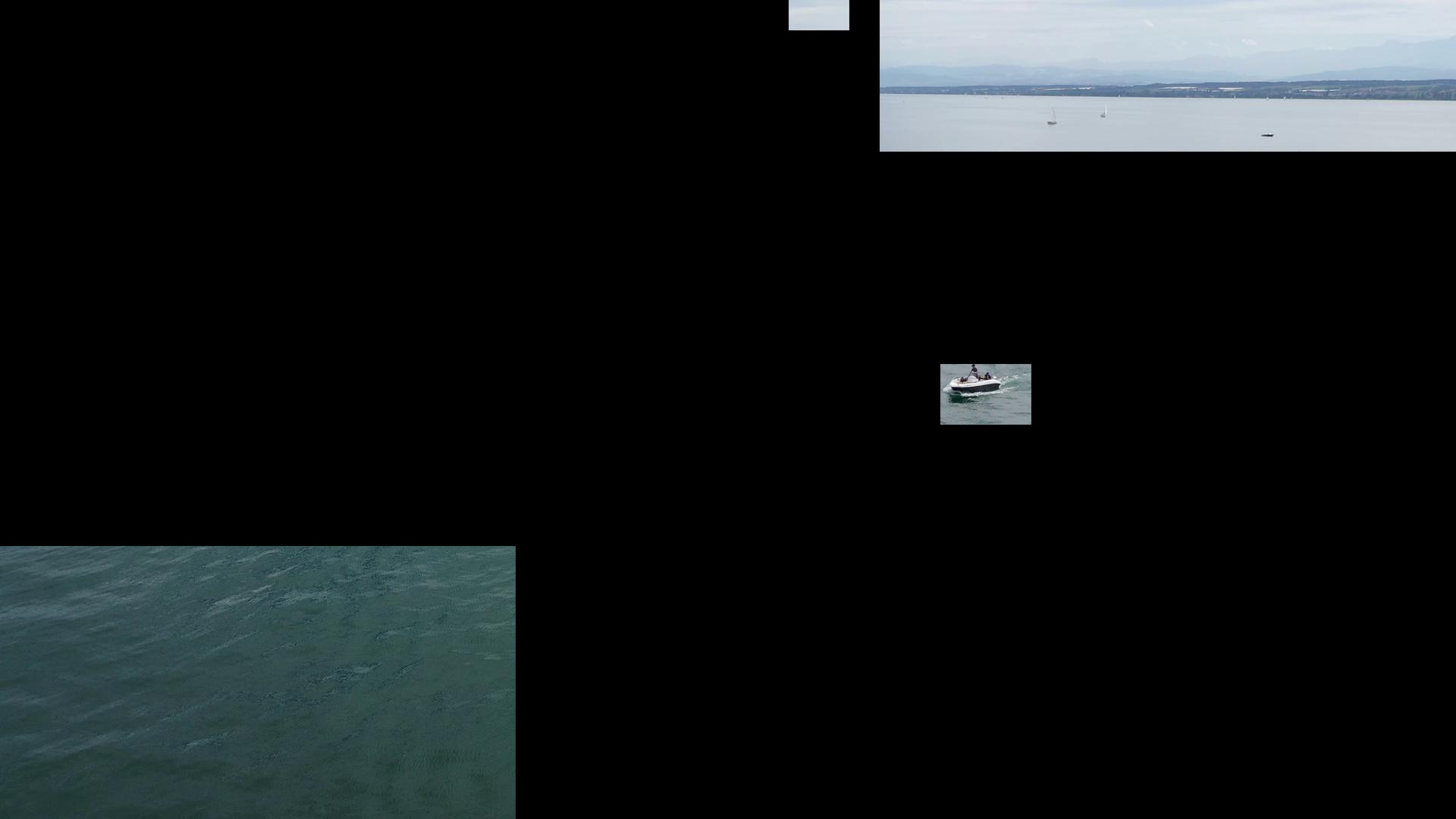}
		
		\includegraphics[trim=0 0 0 0,clip,width=.22\textwidth]{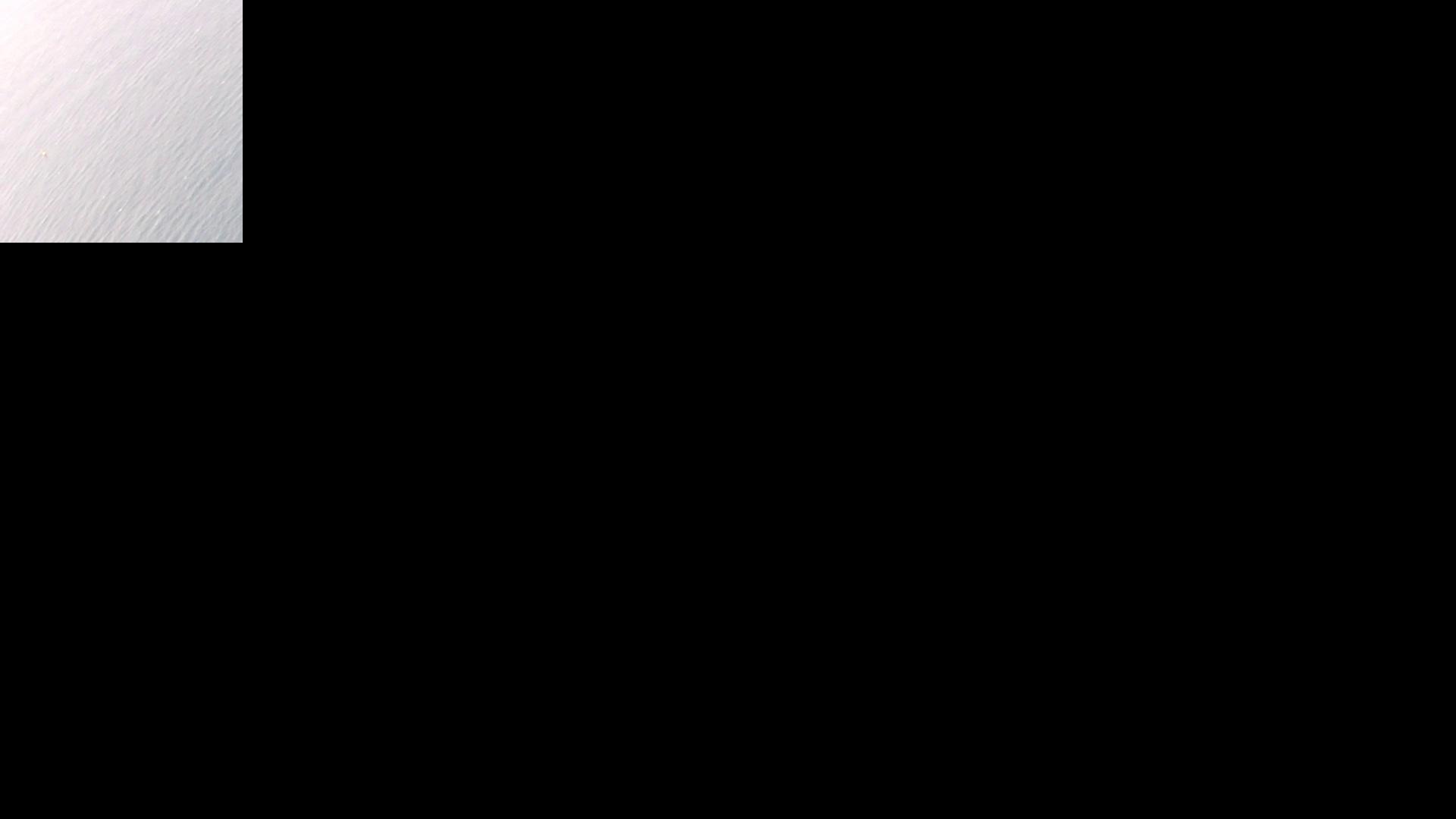}
		
		\includegraphics[trim=0 0 0 0,clip,width=.22\textwidth]{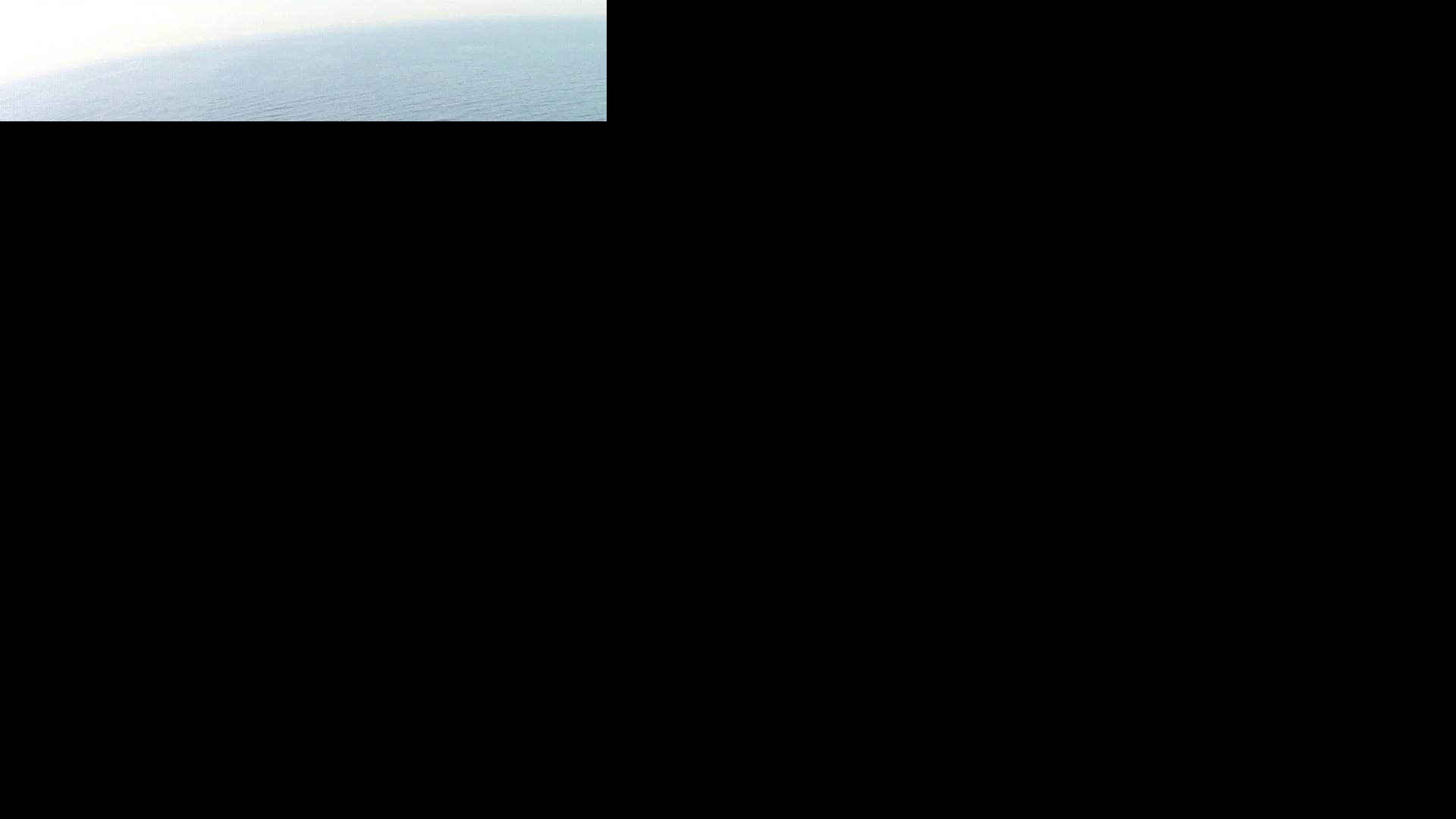}
		
		\\
		
		%\rotatebox{90}{FD err}  \includegraphics[trim=0 0 0 0,clip,width=.22\textwidth]{images/img_Frame_Differencing7.jpg}
		
		%\includegraphics[trim=0 0 0 0,clip,width=.22\textwidth]{images/img_Frame_Differencing1276.jpg}
		
		%\includegraphics[trim=0 0 0 0,clip,width=.22\textwidth]{images/img_Frame_Differencingframe274.jpg}
		
		%\includegraphics[trim=0 0 0 0,clip,width=.22\textwidth]{images/img_Frame_Differencingframe136.jpg}
		
		%\\

		\rotatebox{90}{\ \ FD box}  \includegraphics[trim=0 0 0 0,clip,width=.22\textwidth]{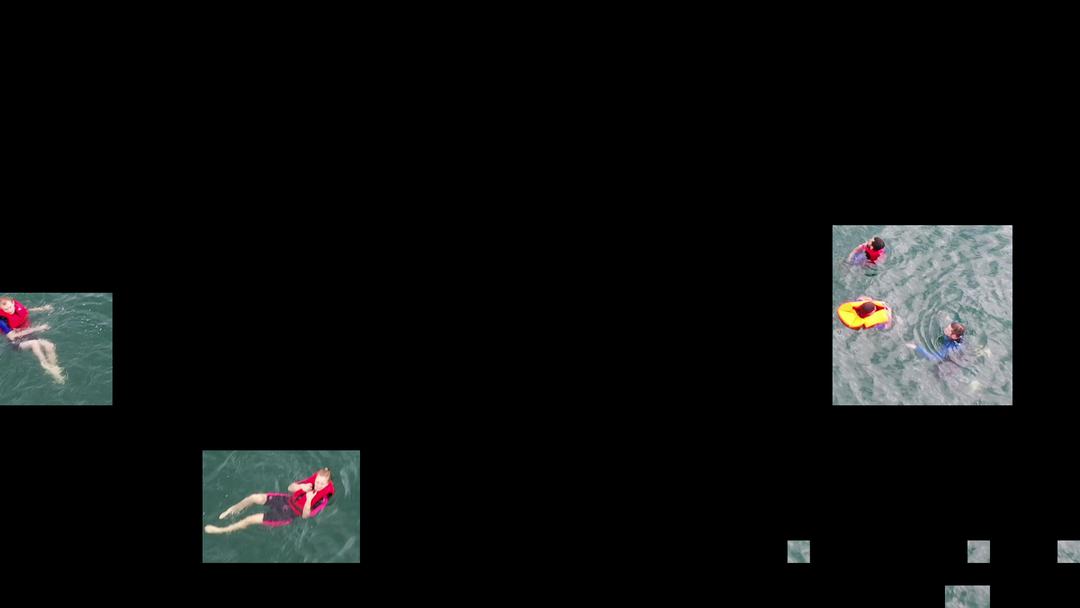}
		
		\includegraphics[trim=0 0 0 0,clip,width=.22\textwidth]{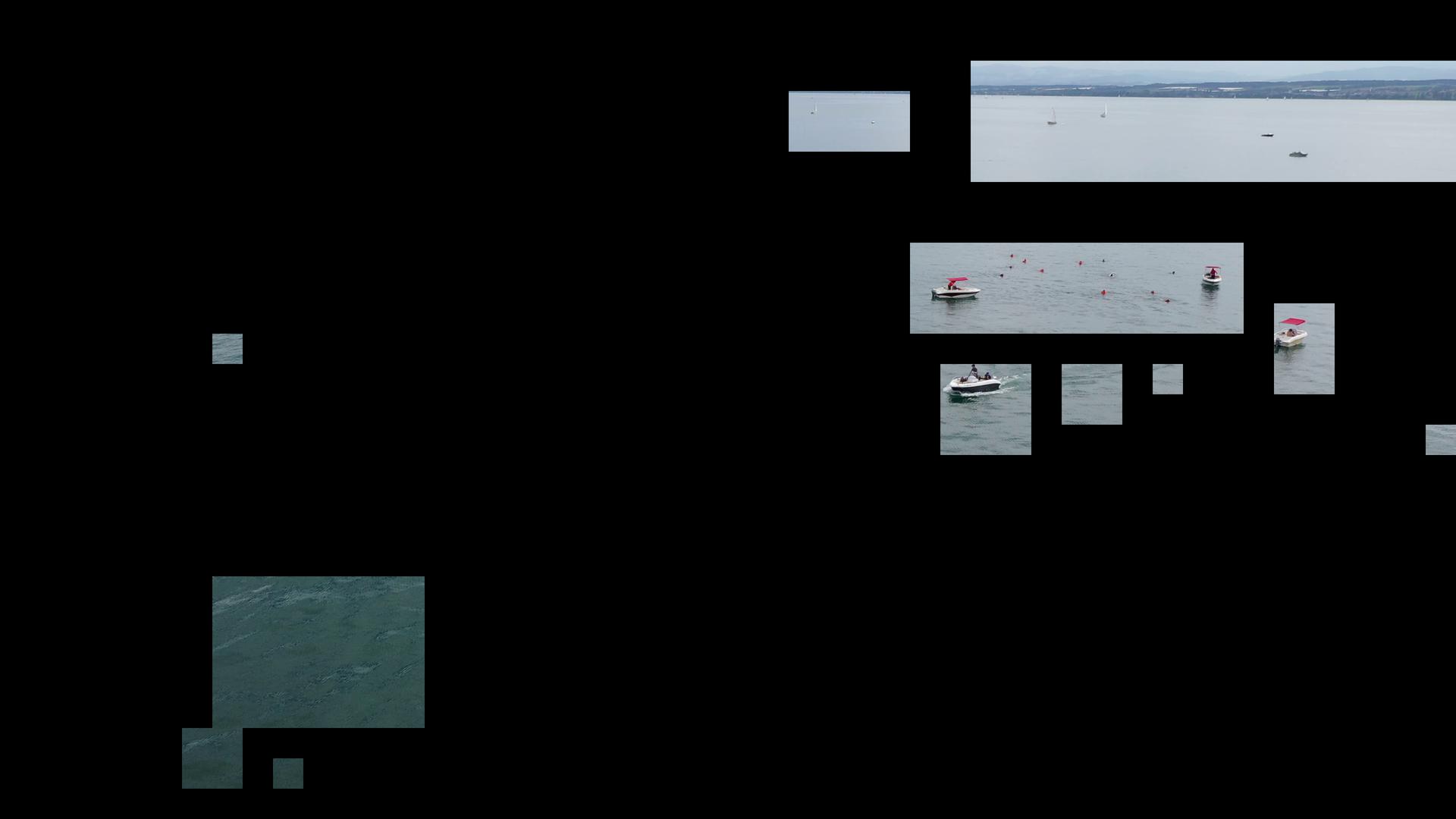}

		\includegraphics[trim=0 0 0 0,clip,width=.22\textwidth]{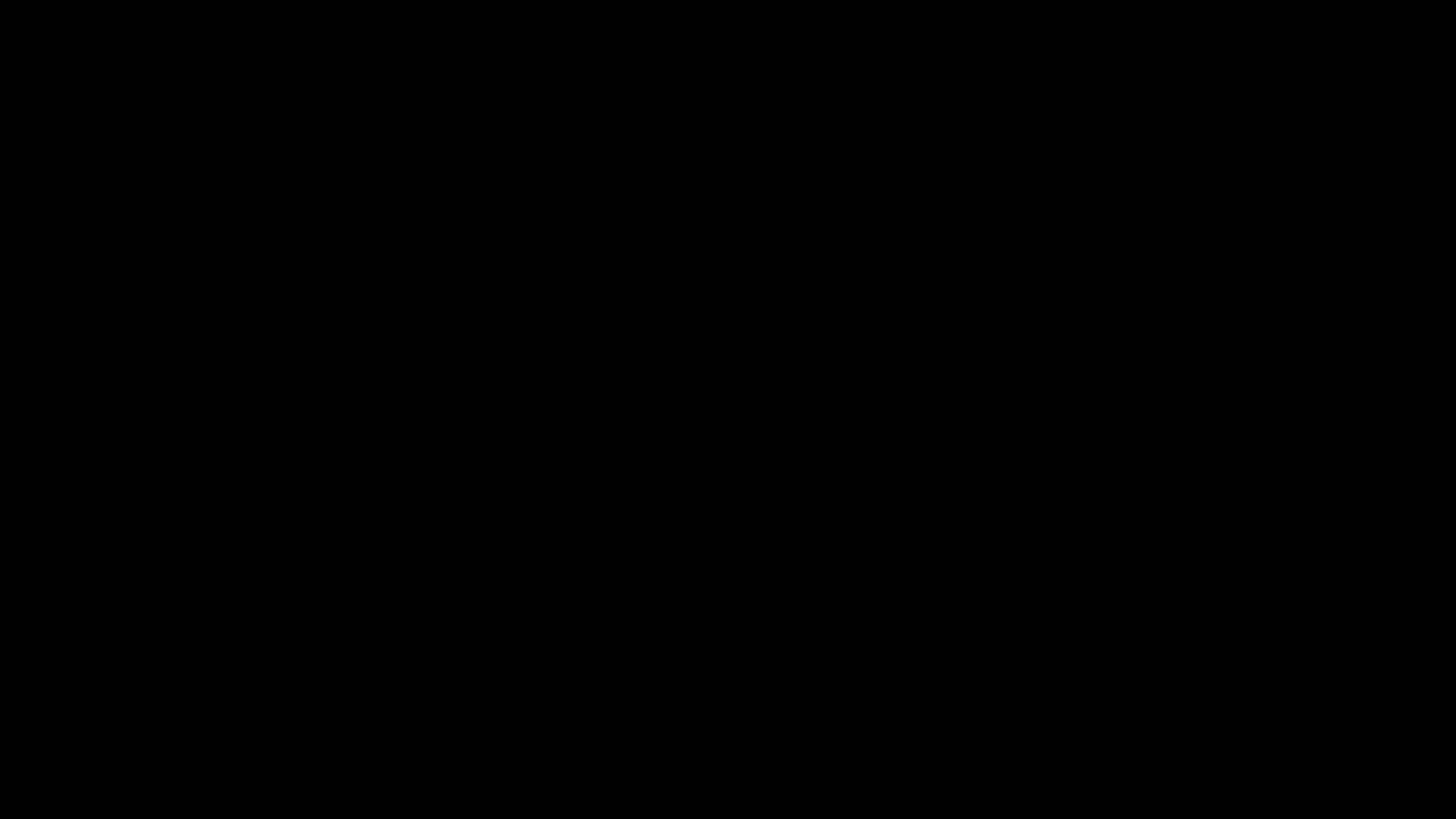}
		
		\includegraphics[trim=0 0 0 0,clip,width=.22\textwidth]{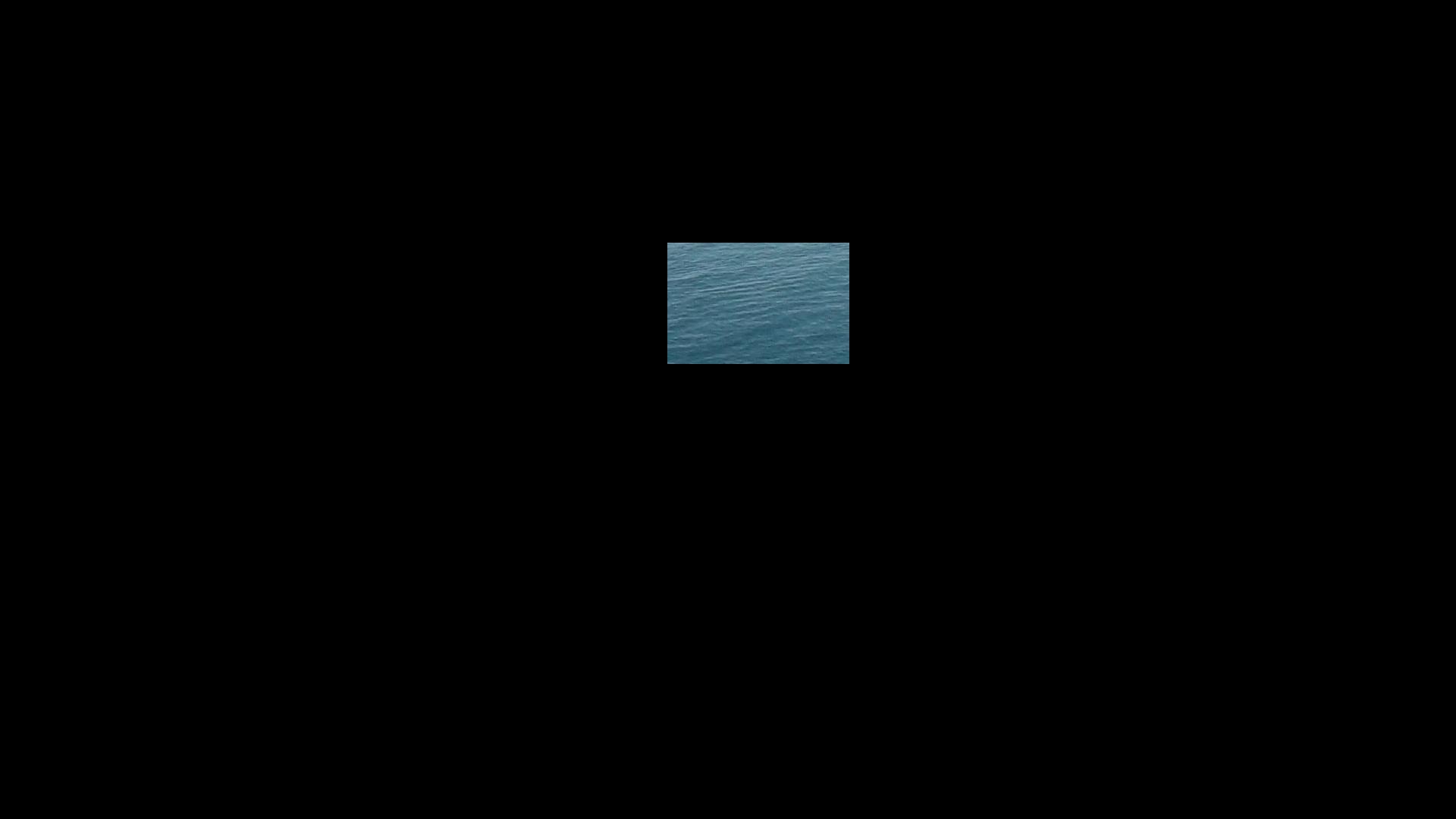}

		\\
        
        %\rotatebox{90}{GMM err}  \includegraphics[trim=0 0 0 0,clip,width=.22\textwidth]{images/img_Gaussian_Mixture5.jpg}
        
         %\includegraphics[trim=0 0 0 0,clip,width=.22\textwidth]{images/img_Gaussian_Mixture1276.jpg}
        
         %\includegraphics[trim=0 0 0 0,clip,width=.22\textwidth]{images/img_Gaussian_Mixtureframe274.jpg}
        
         %\includegraphics[trim=0 0 0 0,clip,width=.22\textwidth]{images/img_Gaussian_Mixtureframe136.jpg}

        %\\
        
        \rotatebox{90}{\ \ GMM box}  \includegraphics[trim=0 0 0 0,clip,width=.22\textwidth]{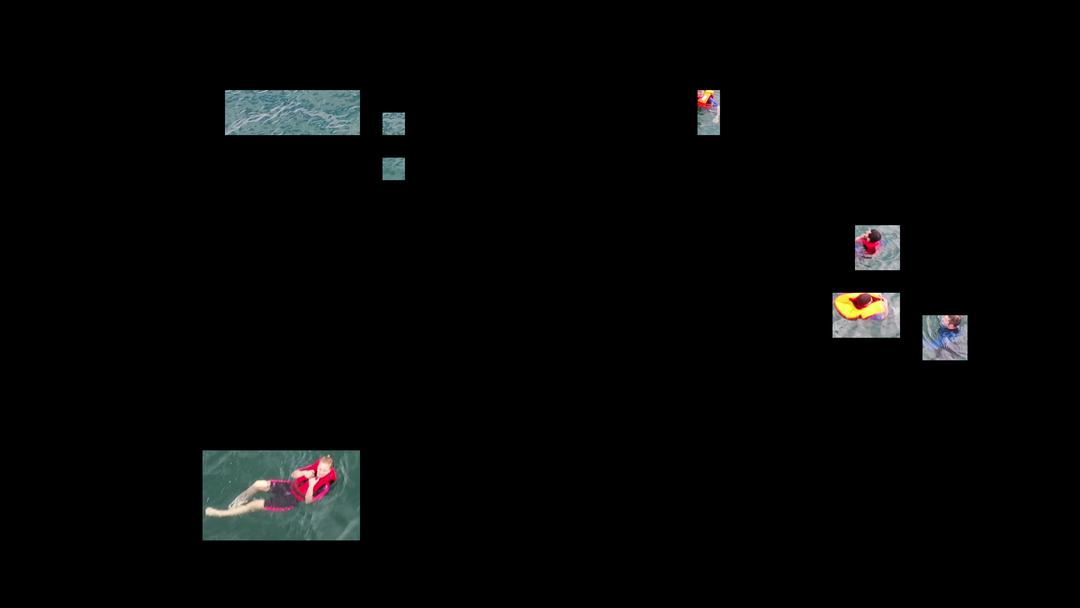}
        
        \includegraphics[trim=0 0 0 0,clip,width=.22\textwidth]{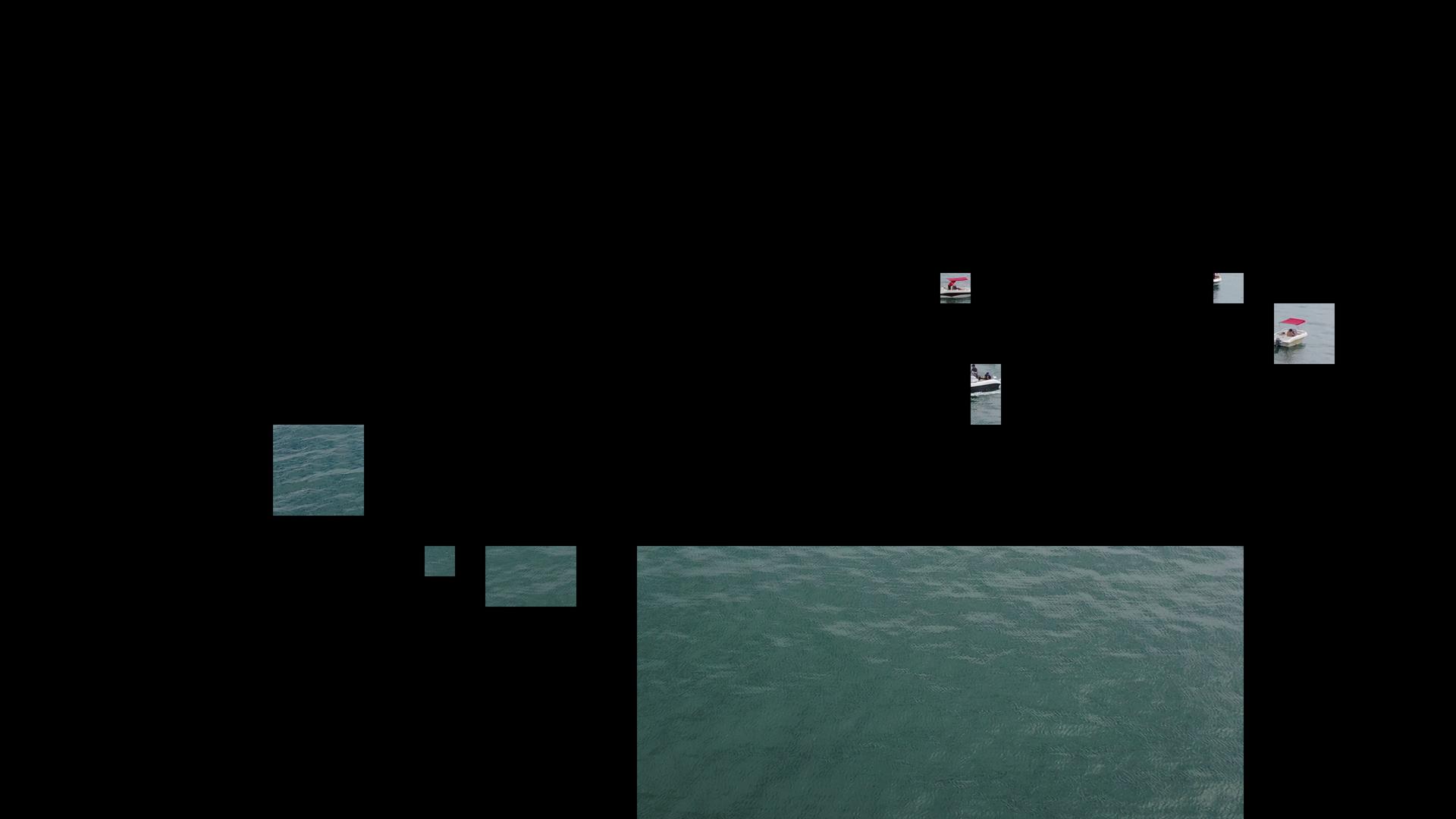}
        
        \includegraphics[trim=0 0 0 0,clip,width=.22\textwidth]{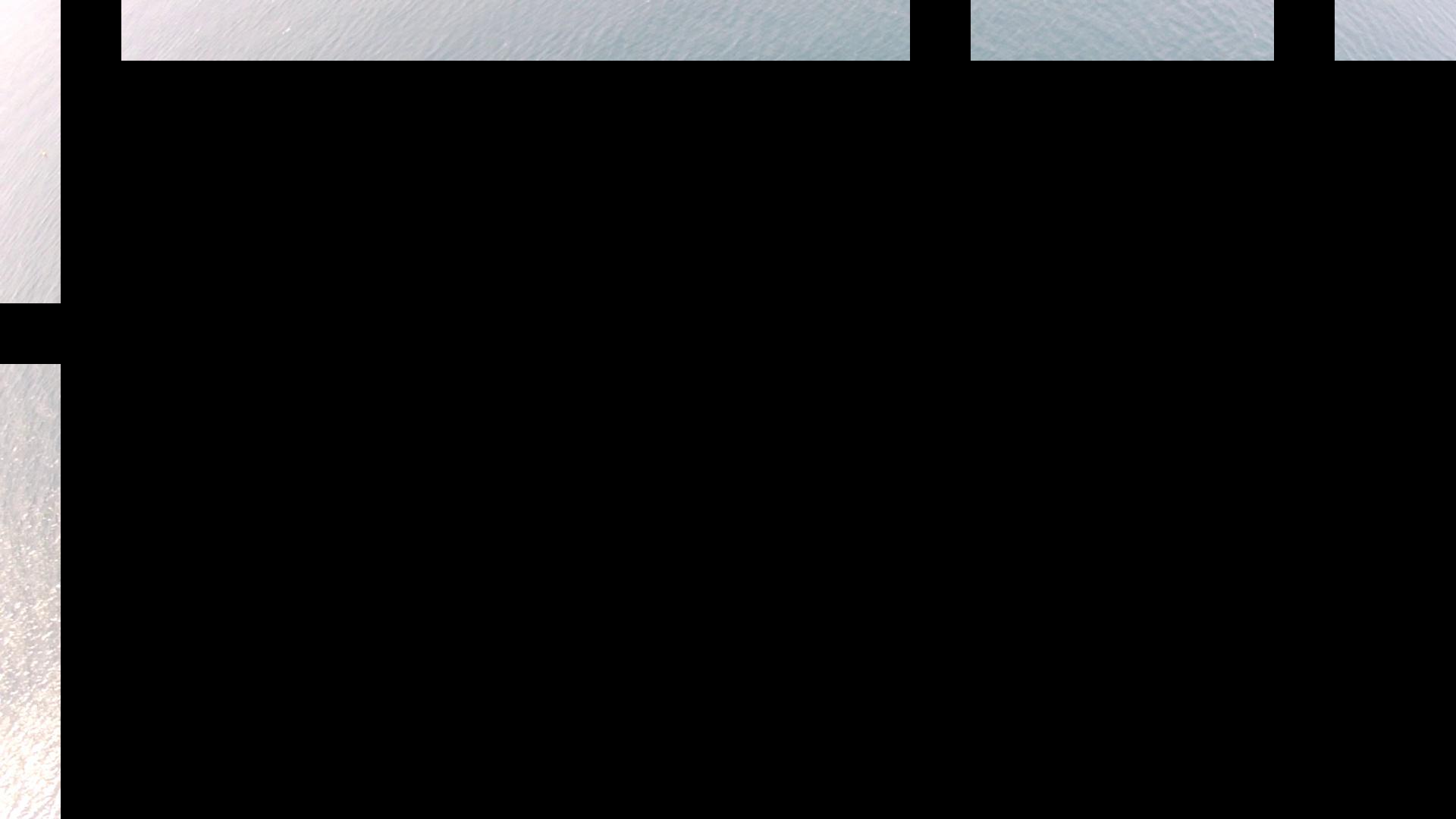}
        
        \includegraphics[trim=0 0 0 0,clip,width=.22\textwidth]{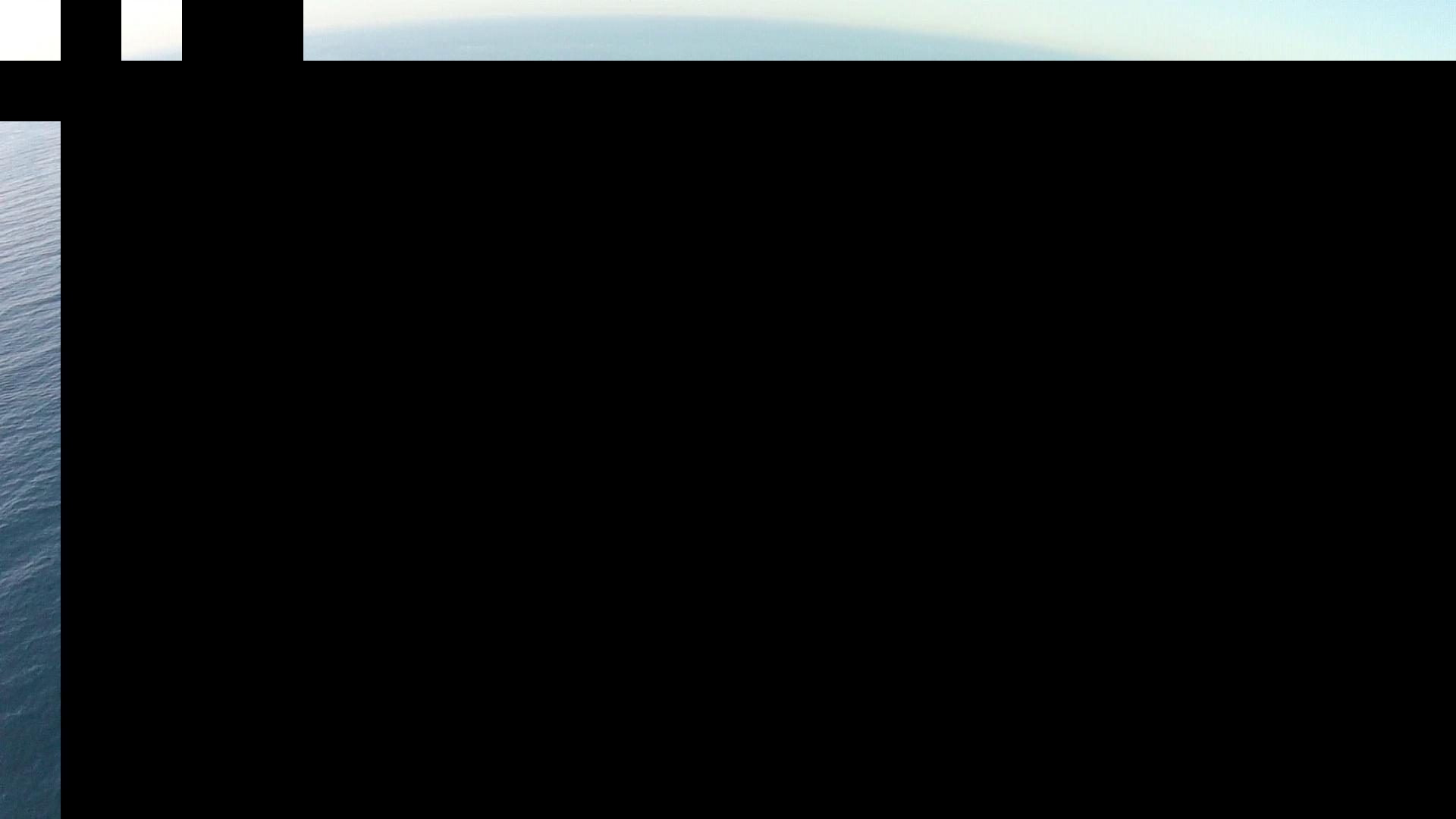}
         
        \\

		\rotatebox{90}{\ \ Auto err}  \includegraphics[trim=0 0 0 0,clip,width=.22\textwidth]{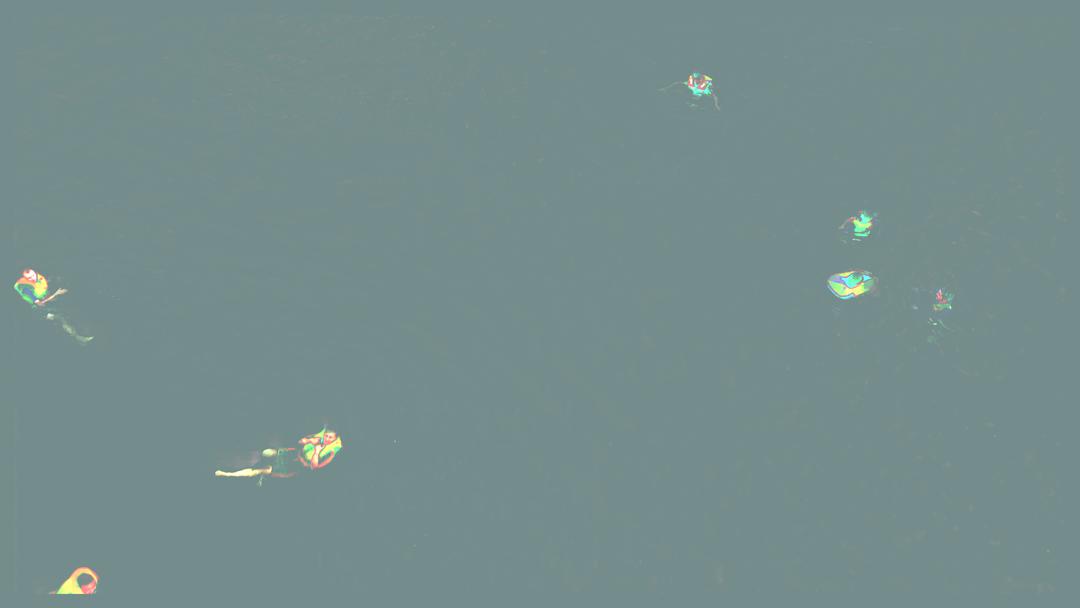}
		
		\includegraphics[trim=0 0 0 0,clip,width=.22\textwidth]{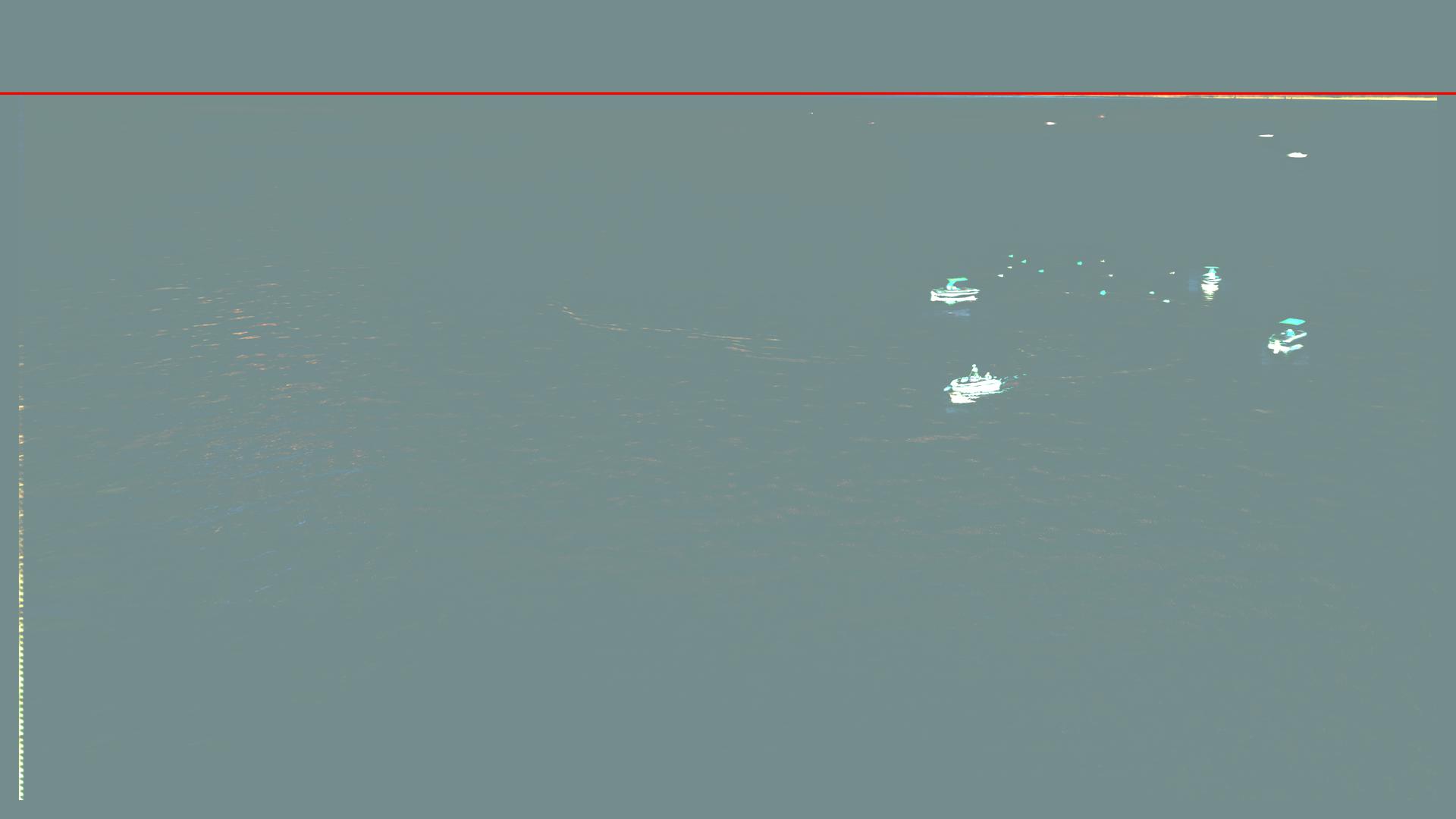}
		
    	 \includegraphics[trim=0 0 0 0,clip,width=.22\textwidth]{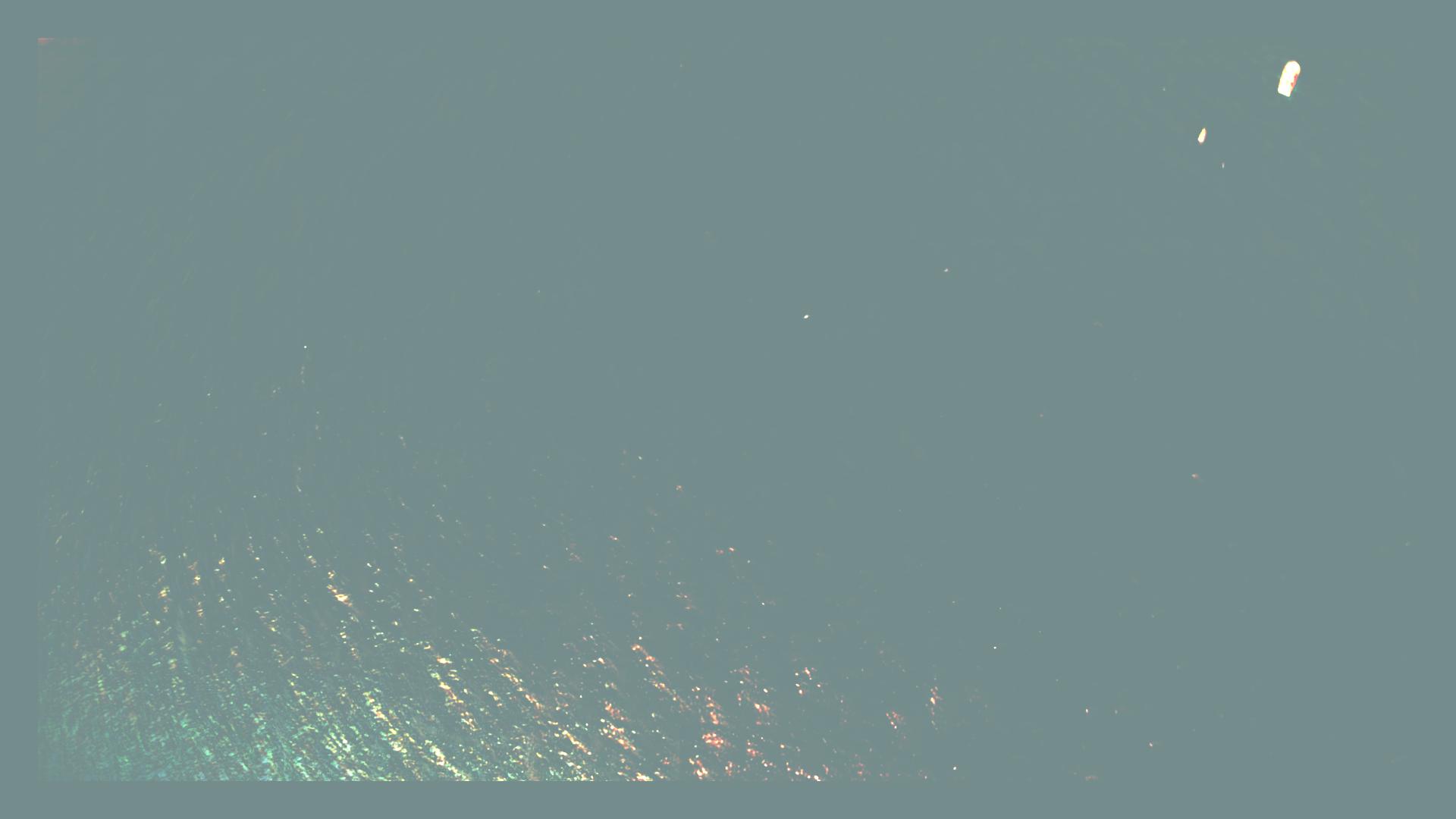}
		
    	 \includegraphics[trim=0 0 0 0,clip,width=.22\textwidth]{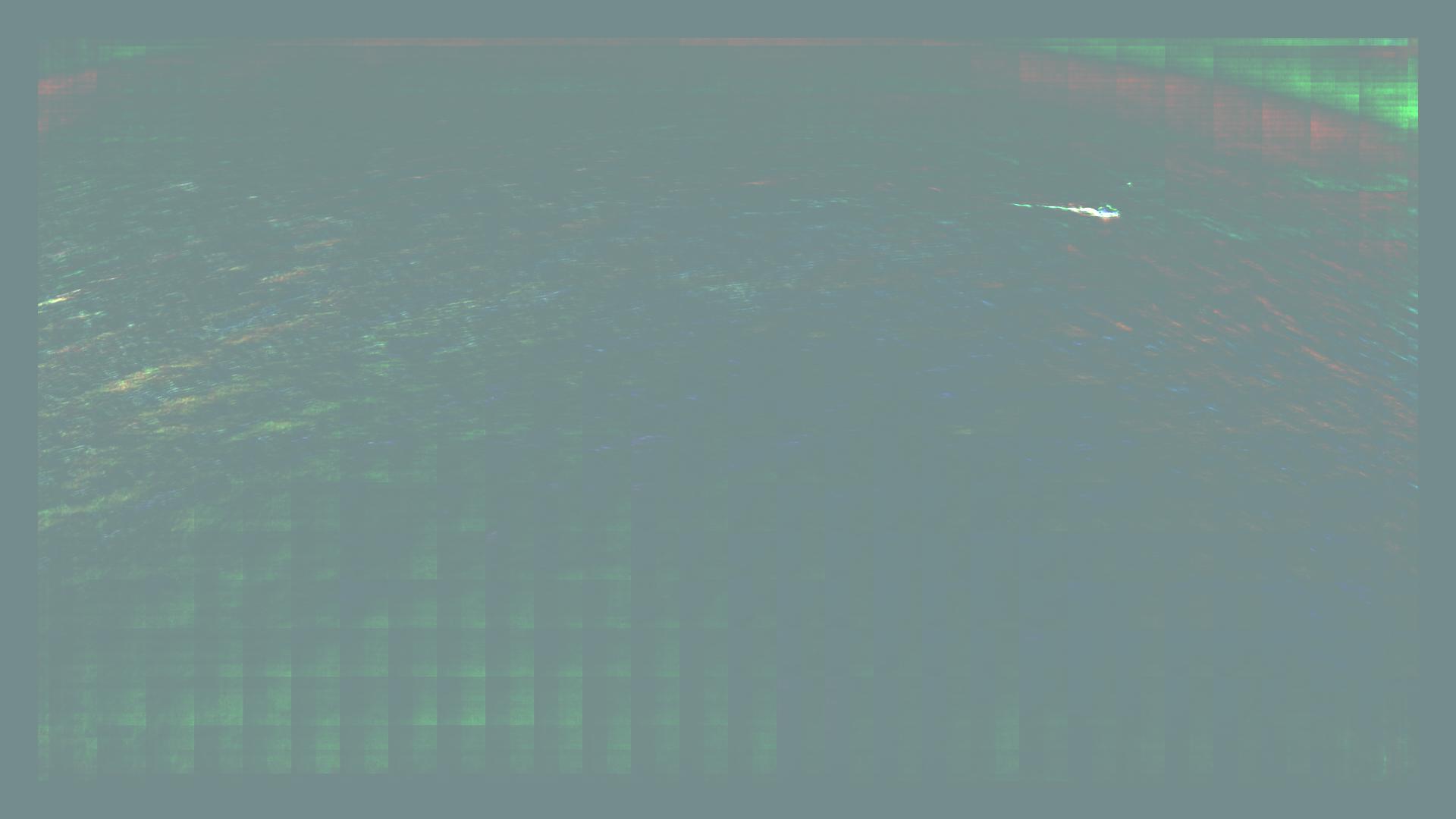}

		\\	
		
		\rotatebox{90}{\ \ Auto box}   \includegraphics[trim=0 0 0 0,clip,width=.22\textwidth]{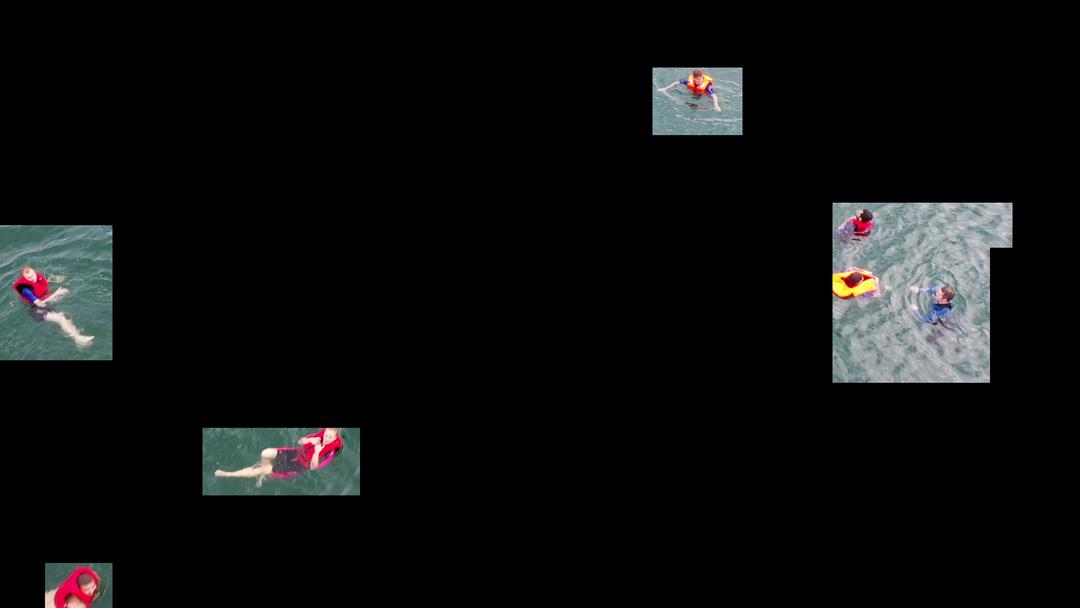}
		
		\includegraphics[trim=0 0 0 0,clip,width=.22\textwidth]{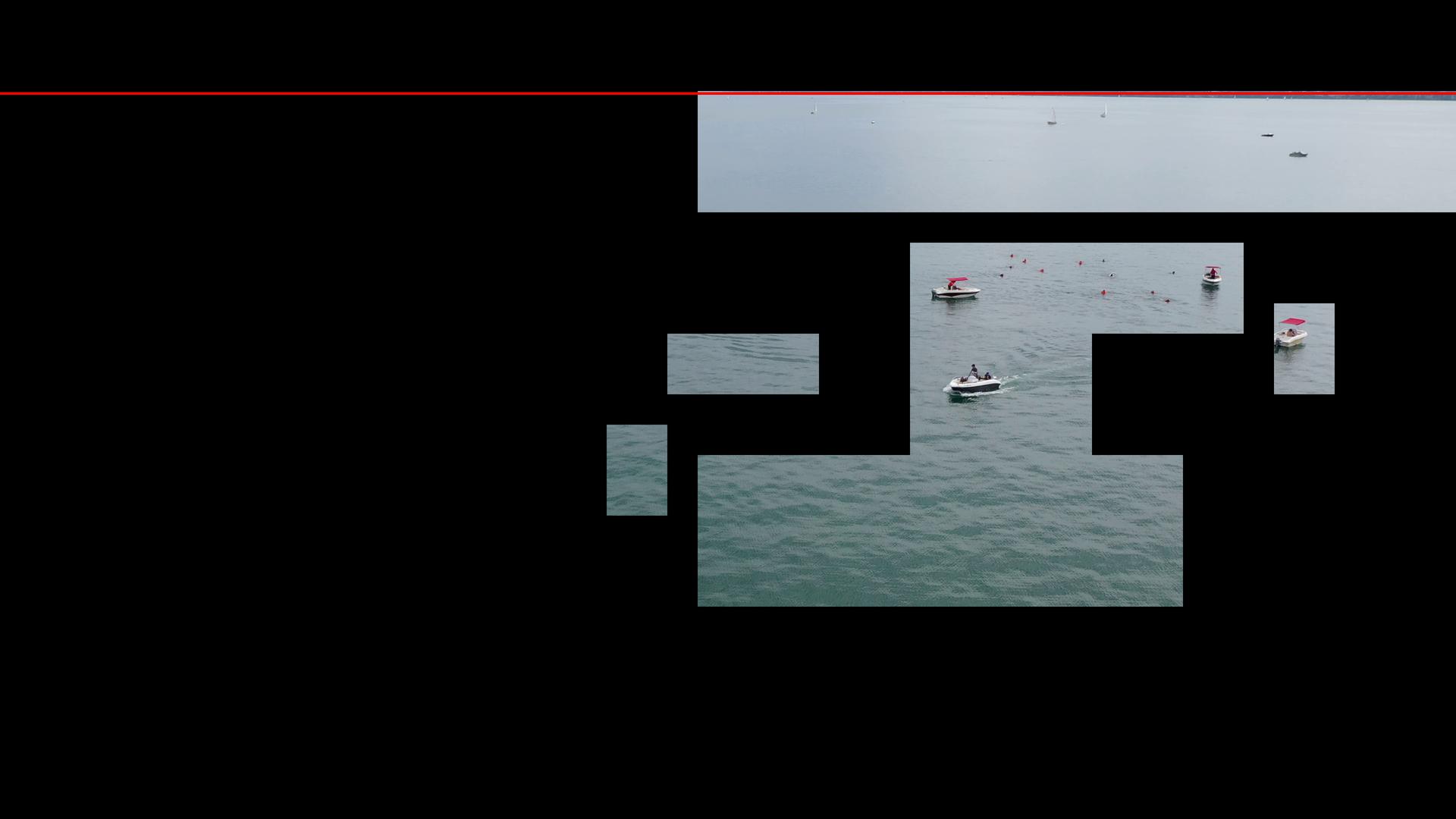}
		
       \includegraphics[trim=0 0 0 0,clip,width=.22\textwidth]{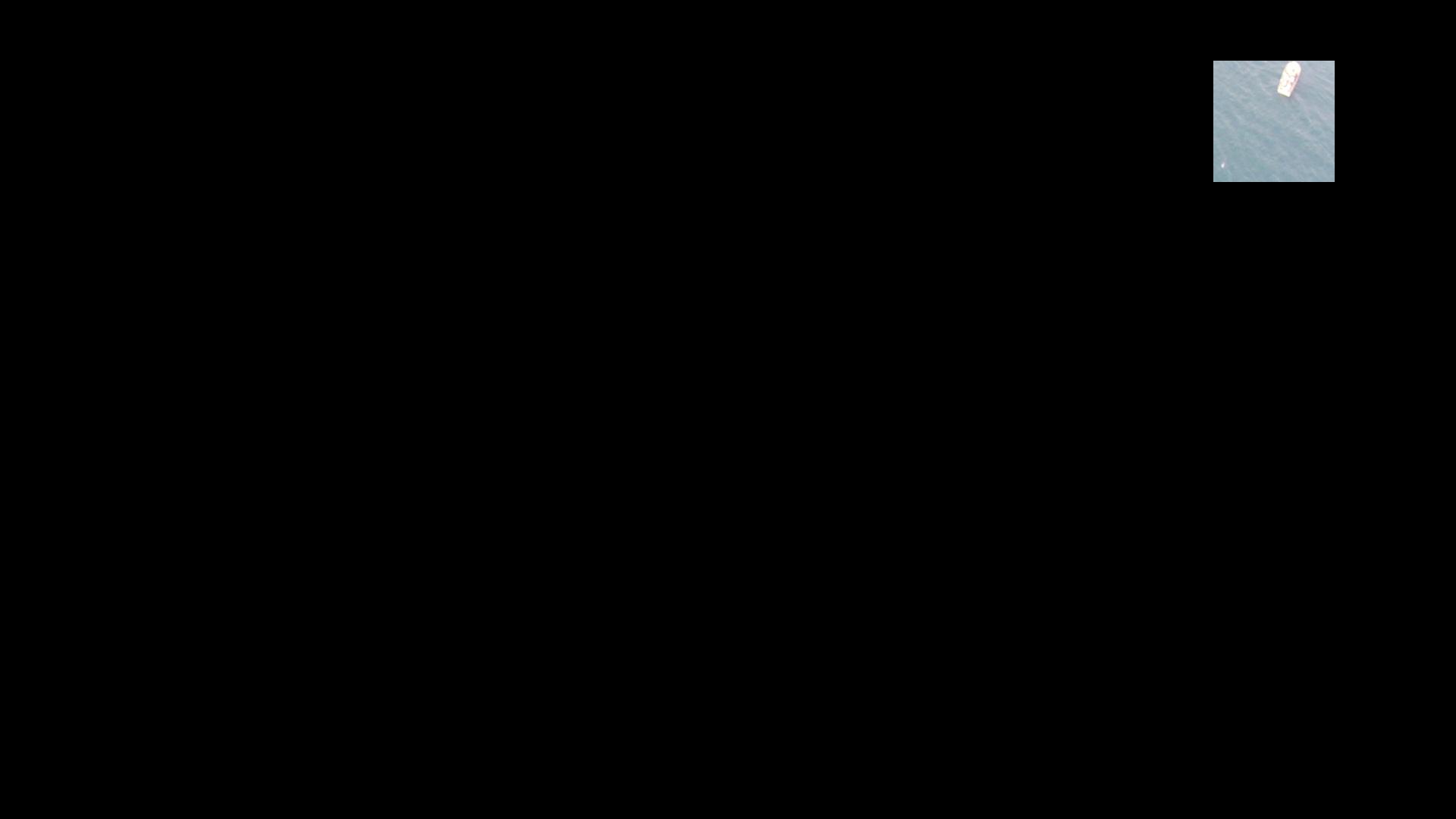}
		
       \includegraphics[trim=0 0 0 0,clip,width=.22\textwidth]{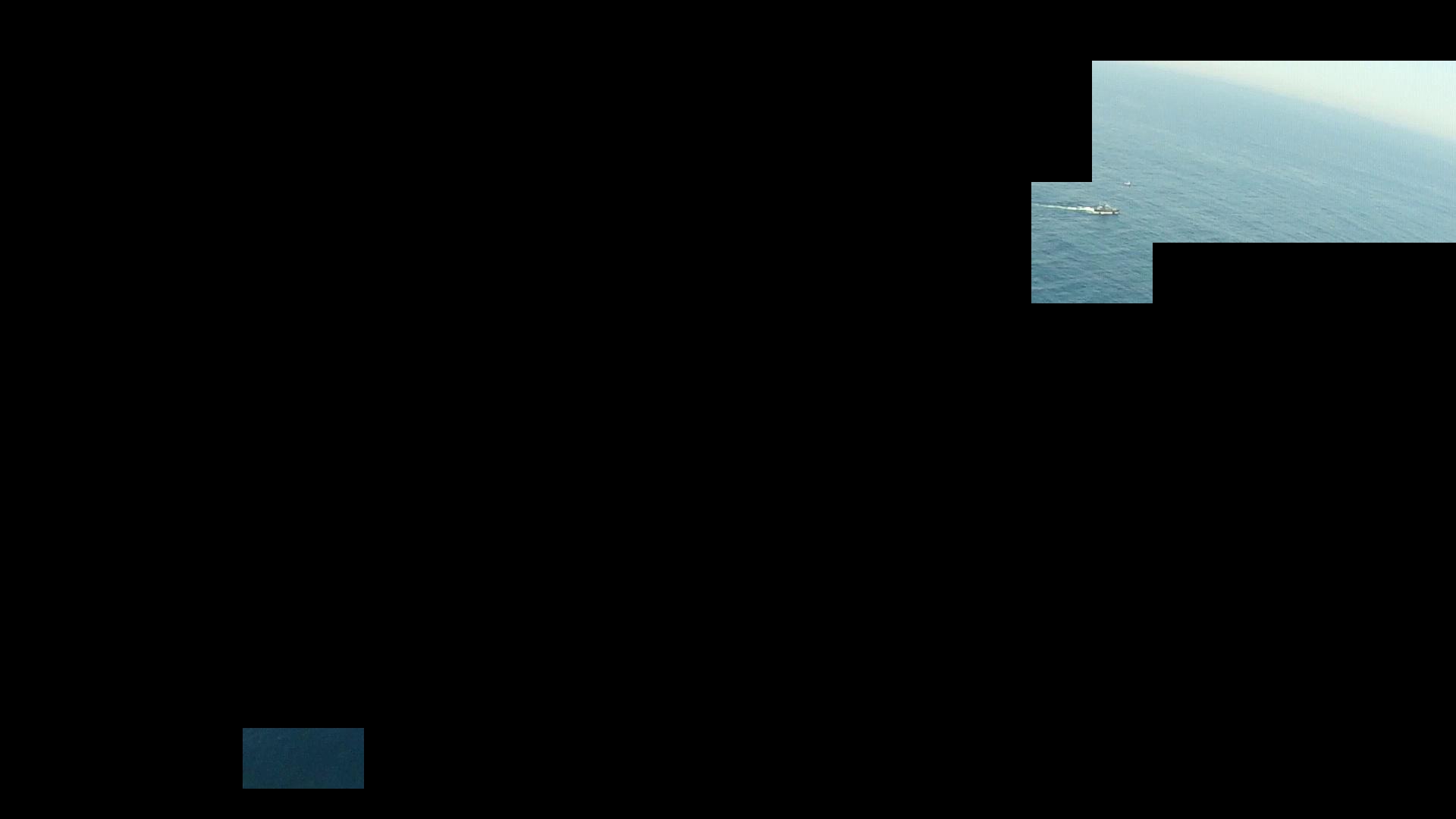}
		\\
		
		\rotatebox{90}{\ \ Recon}  \includegraphics[trim=0 0 0 0,clip,width=.22\textwidth]{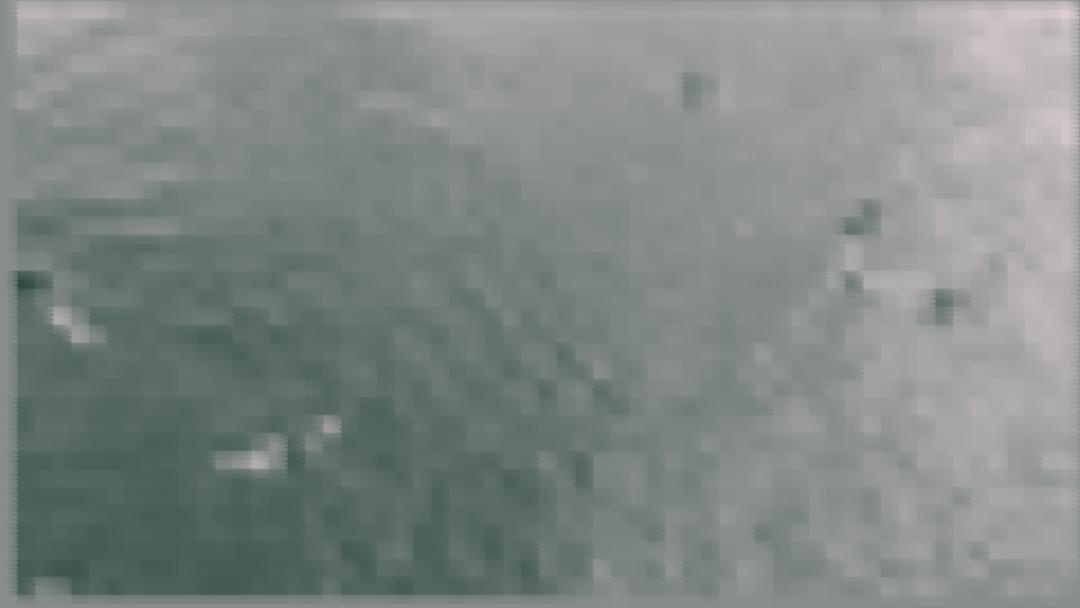}
		
		\includegraphics[trim=0 0 0 0,clip,width=.22\textwidth]{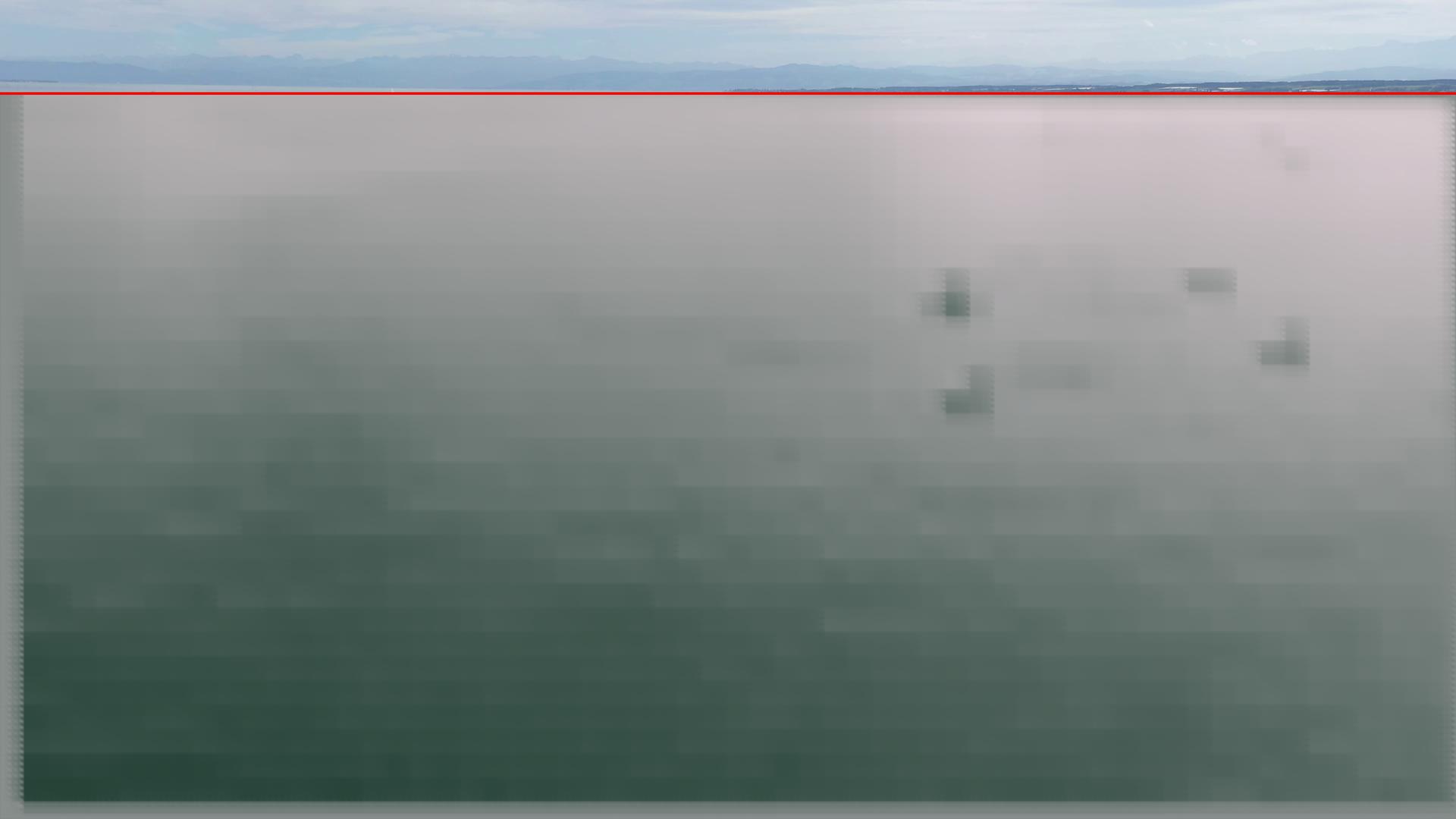}

        \includegraphics[trim=0 0 0 0,clip,width=.22\textwidth]{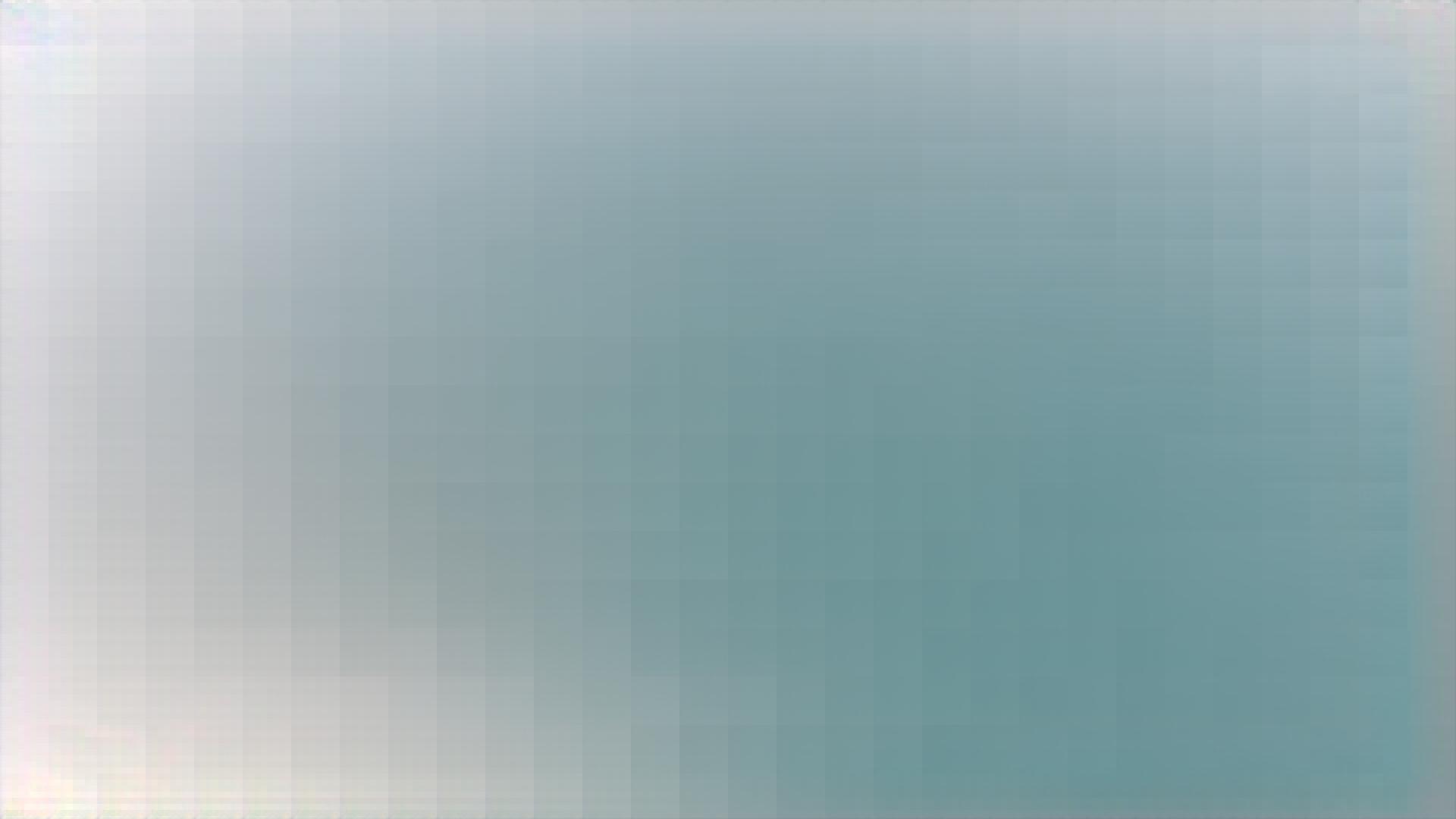}
		
		\includegraphics[trim=0 0 0 0,clip,width=.22\textwidth]{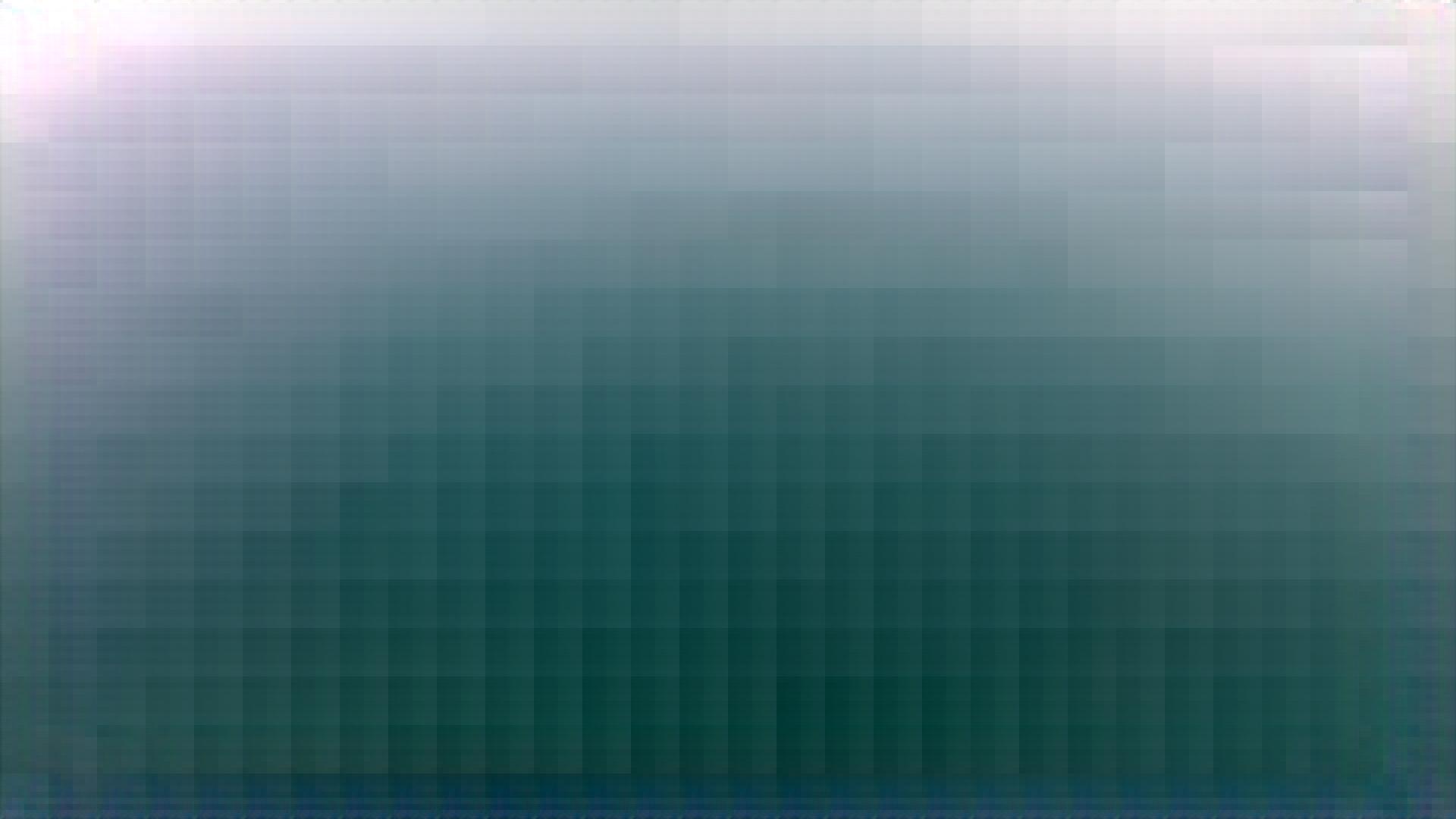}
		
		\\
        
	\end{tabular}
		\caption{Qualitative results for mean filter (MF), frame differencing (FD), Gaussian mixture model (GMM) and the autoencoder (Auto) on SeaDronesSee (left two columns) and Seagull (right two columns). For Auto, we plot the error heat map and the reconstructed image (Recon).}
	\label{fig:manyimgs}
\end{figure*}

As we operate on high resolution videos and in real-time scenarios, we compare to three methods commonly used for background subtraction and anomaly detection: Mean filter (MF) \cite{zhang2012object}, frame differencing (FD) \cite{mohamed2010background} and Gaussian mixture model (GMM) \cite{zivkovic2004improved}. For GMM we use three Gaussians. We extend every method with the grid component and employ the horizon cutter.

%finaldo change maybe citations of these

We evaluate on the following datasets (see also Fig. \ref{fig:manyimgs}):
%finaldo change maybe number of minutes or frames
\begin{itemize}

    \item We use our {\bf Open Water} data set as the training set for SeaDronesSee as the latter does not consist of frames without objects. It consists of $>100000$ frames of open water captured on multiple days with three different 4K video cameras. Similar to SeaDronesSee, we provide precise meta annotations and manually annotate bounding boxes with the same classes as SeaDronesSee for each of the frames.
    
    \item We experiment on the multi-object tracking track of {\bf SeaDronesSee}. It depicts humans, boats and other objects in open water (incl. bboxes) serving as our anomalies. The frames are of 4K resolution and each frame is annotated with precise meta data information. It is challenging since it can contain multiple objects scattered around the frames and due to its diversity in the altitude and angle of view distribution, resulting in different objects' appearances and sizes.
    
    \item The {\bf Seagull} data set features video data showing boats, ships, life rafts and other objects from a fixed wing UAV. It also features video clips containing no objects. The latter serve as our training set for the Seagull test set. The videos are of Full HD resolution and have a  heavy lens distortion and distortion caused by a rolling shutter. See also Figure \ref{fig:manyimgs} for examples.
    
\end{itemize}

We measure the recall given a certain bandwidth (percentage of the video frame transmitted averaged over all frames). Therefore, we consider as evaluation metric $R^{p\%}$, which is the recall over all frames, given that at most $p\%$ of the image may be transmitted. Each region to be transmitted must be encoded by a rectangular bounding box. We consider an object to be correctly detected if there is an overlap with the predicted region of at least 50\%, which is common for aerial object detection \cite{zhu2018vision,varga2021seadronessee,du2018unmanned}. Furthermore, we report the average recall ($AR$) averaged over 10 equidistant percentages $p$ from $p=0.05$ to $p=0.95$, denoted $AR$.

\subsection{Anomaly Detection Performance}

Fig. \ref{fig:arearecall} shows $AR$ and the recall values for all the methods for the 10 transmission percentages $p$, while we interpolate in between. The autoencoder consistently outperforms the baselines for all values of $p$ and for AR. However, the difference is especially visible for low values of $p$, which is the primary use case in this application scenario \cite{steinert2022architecture}. For example, for $p=5\%$ the autoencoder achieves $70.1\%$ and $40.0\%$ recall for SeaDronesSee and Seagull, respectively, which is over $15$ resp. $32$ percent points more than the best baseline. Subsequently, we focus on the case of low $p$.

\begin{table}
\caption{Average $L_1$ recon. error within boxes and outside for $p=5\%$.}

	\begin{tabular}{c|c|c|c|c|c|c}
	&\multicolumn{3}{c|}{SeaDronesSee}&\multicolumn{3}{c}{Seagull}  \\
		 & $err_{b}$ & $err_{r}$ & $\Delta_r$ &$err_{b}$ & $err_{r}$ & $\Delta_r$ \\
		\hline
		MF  & 78.3 & 0.3 & 78.0 & 0.37 & 0.3 & 0.07\\
		FD  & 34.7 & 2.5 & 32.2 & 1.7 & 0.3 &  1.4\\
		GMM  & 4.3 & 0.2 & 4.1 & 0.3 & 0.2 &   0.1\\
		\bf Auto & 79.5 & 0.2 & \bf 79.3 & 2.2 & 0.2 &  \bf 2.0\\ 
	\end{tabular}
	\label{table:errorreconstruction}
\vspace{-5mm}
\end{table}

We report the average reconstruction errors $err_{b}$ (average $L_1$ reconstruction error within ground truth boxes), $err_{r}$ (average $L_1$ reconstruction error rest) and their differences $\Delta_r$ in Table \ref{table:errorreconstruction}. The autoencoder yields higher $\Delta_r$, which shows its ability to discriminate better between normal and anomalous regions. Notably, the values for SeaDronesSee are generally much higher than for Seagull due to Seagull's lower image quality and higher blurriness (see Figure \ref{fig:manyimgs}).

\begin{table}

\caption{Autoencoder ablation experiment on SeaDronesSee. }

  \label{tab:ablation}
      \centering
	\begin{tabular}{c|c|c|c|c}

		Future Frames & -- & \checkmark & \checkmark & \checkmark \\
        Local Noise Remover & -- & -- & \checkmark & \checkmark \\
		Frame Momentum & --  & --  & -- & \checkmark \\ 
		\hline
		$R^{p=5\%}$ & 60.3 & 66.2 & 68.6 & \bf 70.1 \\
	\end{tabular}

\end{table}

%\begin{table}
%\caption{Different components' influences on the autoencoder %performance on SeaDronesSee. }
%\vspace{10mm}
	%\begin{tabular}{c|c|c|c|c}
%
%		F & -- & \checkmark & \checkmark & \checkmark \\
 %       LNR & -- & -- & \checkmark & \checkmark \\
%		F. M & --  & --  & -- & \checkmark \\ 
%		\hline
%		$R^{p=5\%}$ & 60.3 & 66.2 & 68.6 & \bf 70.1 \\
%	\end{tabular}
%	\label{tab:ablation}
%\end{table}

%\begin{table}
%\caption{Average $L_1$ reconstruction error within the regions of interest and outside of them for %$p=5\%$.}
%\begin{center}
%	\begin{tabular}{c|c|c|c|c|c|c|c}
%	&\multicolumn{3}{c|}{SeaDronesSee}&\multicolumn{3}{c|}{Seagull}  \\
%		 & $err_{box}$ & $err_{rest}$ & $\Delta_r$ &$err_{box}$ & $err_{rest}$ & $\Delta_r$ \\
%		\hline
%		\hline
%		MF  & 78.3 & 0.3 & 78.0 & 0.37 & 0.3 & 0.07\\
%		FD  & 34.7 & 2.5 & 32.2 & 1.7 & 0.3 &  1.4\\
%		GMM  & 4.3 & 0.2 & 4.1 & 0.3 & 0.2 &   0.1\\
%		\bf Auto & 79.5 & 0.2 & \bf 79.3 & 2.2 & 0.2 &  \bf 2.0\\ 
%	\end{tabular}
%\end{center}
%\label{table:errorreconstruction}
%\end{table}

%finalDO change order of image with momentum and localnoiseremover

%performance in high movement video clips (maybe for ablation?)
%also frame momentum number for ablation. possibly depending on how fast uav moves

Table \ref{tab:ablation} analyzes the influence of different components. When using future frames, we take the past four frames to predict the fifth. For frame momentum, we use the past two frames. It shows that using future frames yields the greatest benefits. All components improve the performance.

We analyze the influence of the horizon cutter on the performance of the autoencoder. As
only SeaDronesSee incorporates meta data, we perform experiments on this data set. Only 5\% of all frames actually show the horizon. Therefore,
we restrict the influence of the horizon cutter to only that portion, as it does not have any
on the other part. Remarkably, the autoencoder with horizon cutter achieves 86.3\% recall
whereas it only achieves 47.0\% without. As the autoencoder only is trained on frames of
open water, the different image statistics of the sky skew the image reconstruction error on
these parts which expectedly results in a loss in performance. Both experiments used $p = 5$\%.

So far, we considered the case where we only have access to normal frames as training data. However, often we are given some labeled training data. Thus, we propose to use an adversarial training objective where we maximize the prediction penalty of the autoencoder within ground truth boxes and minimize it everywhere else. That way, the model is punished for learning to reconstruct actual anomalies. We evaluate this strategy by comparing it to its naive counterpart, i.e. not backpropagating the loss within boxes.

We compare these two approaches by training on the SeaDronesSee tracking train set and testing on the SeaDronesSee tracking test set. For $p=5\%$,the ignoring yields a recall of $65.7\%$ in contrast to $67.2\%$ with adversarial loss. 

%no negative gradient
%brisk-mountain-388

%current recall of Autoencoder  :  tensor([0.7085])
%current recall of Naive Baseline  :  tensor([0.7472])
%current recall of Frame Differencing  :  tensor([0.7604])
%urrent recall of Gaussian Mixture  :  tensor([0.7121])

%\begin{table}
%\caption{Impacts of different components on speed of Autoencoder.}
%\begin{center}
%\resizebox{10cm}{!}{	
%	\begin{tabular}{c|c|c|c|c|c}
%		komponente1  & komponente2& komponente3 & komponente4& AR & FPS gpus\\
%		\hline
%		\hline
 %%		checkmark & \checkmark\\ 
%
%	\end{tabular}}
%\end{center}

%\label{table:speed_opti}
%\end{table}

\subsection{Obtaining Fewer Bounding Boxes}

Aside from the restriction of choosing at most $p\%$ of the frames, which may be imposed due to a potentially low bandwidth, another restriction may come from common video codecs' inability to process a large number of regions of interest. Therefore, another type of restriction on a region proposer may be the number of regions it yields. 

Thus, we propose to merge regions of interest touching each other at corners using Suzuki's border following method \cite{suzuki1985topological}. As this may yield a larger than allowed area to be transmitted, each resulting box is ranked based on its reconstruction error. This leads to fewer and larger bounding boxes at the expense of a potentially lower recall.

\begin{table}

      \caption{Fewer number of bounding boxes (\#B) via reduced recall on SeaDronesSee.}

      \label{table:fewerbboxes}
      \centering
       \begin{tabular}{c|c|c|c|c}
	& \multicolumn{2}{c|}{Not Merging}&\multicolumn{2}{c}{Merging} \\
		Method  & $\#$B & $R^\text{p=5\%}$ &$\#$B & $R^\text{p=5\%}$   \\ 
		\hline
		MF \cite{zhang2012object} & \bf 65 & 54.8 & 7  &  49.6\\
		FD \cite{mohamed2010background} & \bf 65 & 55.0 & 8 & 53.2\\
		GMM \cite{zivkovic2004improved} & \bf 65 & 0.2 & \bf 5 & 0.1\\
		\bf Auto & \bf 65 & \bf 70.1 & 6 & \bf 64.3\\
	\end{tabular}

\end{table}

Table \ref{table:fewerbboxes} shows the number of boxes \#B and the recall for the standard and the merging method for $p=5\%$. Note that without merging we have the same number of bounding boxes for all the methods since we allow $5\%$ of the area of the image to be transmitted. We can substantially decrease the number of bounding boxes at the cost of a slightly lower recall. We note that this also highly depends on the anomaly distribution since for clustered anomalies it is easier to merge bounding boxes (see Fig. \ref{fig:manyimgs}).

%\begin{table}
%\caption{With the use of a bounding box merging method, we can decrease the number of bounding boxes at %the cost of a slightly reduced recall. Experiment on SeaDronesSee.}
%\begin{center}
%	\begin{tabular}{c|c|c|c|c}
%	& \multicolumn{2}{c|}{Not Merging}&\multicolumn{2}{c}{Merging} \\
%		Method  & $\# B$ & $R^\text{p=5\%}$ &$\# B$ & $R^\text{p=5\%}$   \\ 
%		\hline
%		\hline
%		MF  & \bf 65 & 54.8 & 7  &  49.6\\
%		FD & \bf 65 & 55.0 & 8 & 53.2\\
%		GMM  & \bf 65 & 0.2 & \bf 5 & 0.1\\
%		\bf Auto & \bf 65 & \bf 70.1 & 6 & \bf 64.3\\
%	\end{tabular}
%\end{center}
%\label{table:fewerbboxes}
%\end{table}

%\begin{wrapfigure}{r}{0.5\textwidth}
%   \centering
% \caption{NVIDIA Xavier speed comparison for 4K resolution (SeaDronesSee) and 1K (Seagull) reported in frames per second.}
%   \begin{tabular}{ccc}
%   Method & 4K & 1K \\
%   \hline
%   \hline
%    MF & 50 & ?? \\
%    FD & 62 & ??\\
%    GMM & 17 & ??\\
%    Auto & 27 & ??\\
%   \end{tabular}

% \label{fig:speeds}

%\end{wrapfigure}

%\subsection{Transfer to Seagull or other?}

%Gmm sensitive to image vibration/movement and skewness am rnad  bze belichtung. deshalb %nur am rand sachen erkannt

\subsection{Running Times}

Finally, we consider the running times of the individual methods on embedded hardware. We deploy them on an NVIDIA Xavier \cite{nvidiaxavier} mounted on a DJI Matrice 100. We transform all methods into optimized engines using TensorRT \cite{vanholder2016efficient} and set the Xavier to MAX-N mode and report the running times averaged over 1000 frames. Table \ref{table:fpstable} shows the speed comparison between traditional and modern (U-NET \cite{liu2018future}, CFLOW \cite{gudovskiy2022cflow}) methods. The much simpler baselines run in real-time, while the modern methods are slow.

For completeness, we replaced our architecture with the popular UNet architecture and trained it on SeaDronesSee using halved resolution and filter dimensions (more did not fit into a 3090Ti w/ $24$GB). Interestingly, the performance trailed the performance of our method (78.1 AR). This led
us to the conjecture that the high resolution is crucial in
this application, which makes sense if we consider that
many objects are of $\approx 20$px size.

\begin{table}

      \caption{Running times in FPS. Bold values depict real-time methods.}

      \label{table:fpstable}
      \centering
       \begin{tabular}{c|c|c|c|c|c|c}
         & MF  & FD & GMM & U-NET & CFLOW & Auto\\
		\hline
	1K  & \bf{64} & \bf{70} & \bf{35}  & \textcolor{red}{8} & \textcolor{red}{12} & \bf{48}  \\
		4K & \bf{50}& \bf{62} & \textcolor{red}{17} & \textcolor{red}{1}& \textcolor{red}{3}& \bf{27}\\

	\end{tabular}

\end{table}

%appendixDO subsection failure cases?

\section{Conclusion and Outlook}

We formulated the novel problem in maritime SaR of finding relevant regions of interest in a low-resource real-time and high-resolution scenario.
We show that an autoencoder-based future frame prediction model is a promising direction even in a resource constrained setting. The benchmark is publicly available and we hope that the field of maritime SaR will be advanced by means of fast neural networks in the future.

\newpage

\bibliographystyle{IEEEtran}
\bibliography{IEEEabrv,IEEEexample}

\end{document}